%% file: main.tex
\documentclass[11pt]{article}
\usepackage{pdfpages}
\usepackage[preprint]{acl}
\usepackage{float}
\usepackage{times}
\usepackage{latexsym}
\usepackage[T1]{fontenc}
\usepackage[utf8]{inputenc}
\usepackage{microtype}
\usepackage{inconsolata}
\usepackage{graphicx}
\usepackage{booktabs}
\usepackage[table]{xcolor} %
\usepackage{float}
\usepackage{subcaption}

\usepackage{amsmath, amssymb}
\usepackage{algorithm}
\usepackage{algorithmic}
\usepackage{xcolor}
\usepackage{tabularx}
\usepackage{pifont}
\newcommand{\cmark}{\ding{51}}
\newcommand{\xmark}{\ding{55}}

\usepackage{multirow}   
\usepackage{stfloats}
\usepackage[most]{tcolorbox}
\tcbuselibrary{skins, breakable}
\definecolor{darkgreen}{RGB}{0,128,0}
\definecolor{githubicon}{HTML}{24292F}
\definecolor{projecticon}{HTML}{00A98F}
\tcbuselibrary{listings,breakable,skins}
\usepackage{xcolor}
\usepackage{listings}
\usepackage{fontawesome5}
\definecolor{promptbg}{HTML}{FFF2C6}
\definecolor{promptbar}{HTML}{C99A00}
\definecolor{promptframe}{HTML}{D6A400}

\lstdefinestyle{promptstyle}{
    basicstyle=\ttfamily\scriptsize,
    columns=fullflexible,
    keepspaces=true,
    showstringspaces=false,
    breaklines=true,
    breakatwhitespace=false,
    tabsize=2
}

\newtcblisting{promptbox}[1]{
    enhanced,
    breakable,
    listing only,
    title={#1},
    colback=promptbg,
    colframe=promptframe,
    coltitle=white,
    colbacktitle=promptbar,
    fonttitle=\bfseries\ttfamily\small,
    arc=2mm,
    boxrule=0.5pt,
    left=1mm,
    right=1mm,
    top=1mm,
    bottom=1mm,
    listing options={style=promptstyle}
}

\tcbuselibrary{listings,breakable,skins}
\usepackage{xcolor}
\usepackage{listings}

\definecolor{caseblue}{HTML}{EAF3FF}
\definecolor{caseblueframe}{HTML}{4E8DD6}
\definecolor{caseyellow}{HTML}{FFF4CE}
\definecolor{caseyellowframe}{HTML}{D7A600}
\definecolor{casegreen}{HTML}{EAF8EA}
\definecolor{casegreenframe}{HTML}{2DA44E}
\definecolor{casered}{HTML}{FFF0F0}
\definecolor{caseredframe}{HTML}{D1242F}

\newtcolorbox{tracebox}[1]{
  enhanced, breakable,
  title={#1},
  colback=casered,
  colframe=caseredframe,
  colbacktitle=caseredframe,
  coltitle=white,
  fonttitle=\bfseries,
  boxrule=0.6pt,
  arc=1.5mm,
  left=1mm,right=1mm,top=1mm,bottom=1mm,
  before skip=0.5em,
  after skip=0.8em
}

\newtcolorbox{criticbox}[1]{
  enhanced, breakable,
  title={#1},
  colback=caseyellow,
  colframe=caseyellowframe,
  colbacktitle=caseyellowframe,
  coltitle=black,
  fonttitle=\bfseries,
  boxrule=0.6pt,
  arc=1.5mm,
  left=1mm,right=1mm,top=1mm,bottom=1mm,
  before skip=0.5em,
  after skip=0.8em
}

\newtcblisting{skilldiffbox}[1]{
  enhanced, breakable,
  listing only,
  title={#1},
  colback=caseblue,
  colframe=caseblueframe,
  colbacktitle=caseblueframe,
  coltitle=white,
  fonttitle=\bfseries,
  boxrule=0.6pt,
  arc=1.5mm,
  left=1mm,right=1mm,top=1mm,bottom=1mm,
  before skip=0.5em,
  after skip=0.8em,
  listing options={
    basicstyle=\ttfamily\scriptsize,
    columns=fullflexible,
    keepspaces=true,
    breaklines=true,
    showstringspaces=false
  }
}

\newtcolorbox{successbox}[1]{
  enhanced, breakable,
  title={#1},
  colback=casegreen,
  colframe=casegreenframe,
  colbacktitle=casegreenframe,
  coltitle=white,
  fonttitle=\bfseries,
  boxrule=0.6pt,
  arc=1.5mm,
  left=1mm,right=1mm,top=1mm,bottom=1mm,
  before skip=0.5em,
  after skip=0.8em
}

\makeatletter
\renewcommand{\subsubsection}{\@startsection{subsubsection}{3}{\z@}%
  {-2ex plus -0.5ex minus -0.2ex}%
  {0.8ex plus 0.2ex}%
  {\normalfont\small\bfseries}}  %
\makeatother

\definecolor{myblue}{RGB}{30, 144, 255}

\title{Reflect, Revise, Reuse: Training-Free Skill Evolution for GUI Agents}

\author{
  Bofan Chen\textsuperscript{1}\thanks{\,Equal contribution.} \quad
  Boxuan Zhang\textsuperscript{1,2}\footnotemark[1] \quad
  Fei Tang\textsuperscript{1} \quad
  Zhengxi Lu\textsuperscript{1} \quad
  Yong Du\textsuperscript{1} \\
  \bfseries
  Tongbo Chen\textsuperscript{1} \quad
  Weiming Lu\textsuperscript{1} \quad
  Jun Xiao\textsuperscript{1} \quad
  Yueting Zhuang\textsuperscript{1} \quad
  Yongliang Shen\textsuperscript{1}\thanks{\,Corresponding author.} \\[2pt]
  \mdseries
  \textsuperscript{1}Zhejiang University \quad
  \textsuperscript{2}UESTC \\
  \texttt{\{syl\}@zju.edu.cn} \\[3pt]
  \href{https://github.com/ZJU-REAL/EvoSkill-GUI}{%
    \textcolor{githubicon}{\faGithub}\,\texttt{Code}}
  \quad
  \href{https://zju-real.github.io/EvoSkill-GUI/}{%
    \textcolor{projecticon}{\faGlobe}\,\texttt{Project Page}}
}

\begin{document}
\maketitle

\begin{abstract}

GUI agents execute long-horizon tasks on dynamic graphical user interfaces, where pop-ups, delayed loads, and relocated widgets routinely invalidate plans fixed before execution. Recent agent-skill frameworks encapsulate reusable procedural knowledge to mitigate this, yet existing skill designs are largely developed without targeting GUI execution dynamics and treat skills as static artifacts produced before deployment rather than living procedural knowledge that improves through it. We argue that what GUI agents need is not better static skills, but skills that can be revised from execution feedback at deployment time, without additional training. We propose \textbf{EvoSkill-GUI}, a training-free framework in which each skill is a structured multi-file package containing retrieval metadata, executable plans, backup localization, failure-recovery rules, accessibility utilities, and failure cases. EvoSkill-GUI operates through a \textbf{\emph{reflect-revise-reuse}} loop: the executor performs instant in-rollout revisions, an isolated critic diagnoses failed trajectories under strict information isolation, and the executor edits specific skill files through a restricted tool interface. Across MobileWorld, AndroidWorld, and OSWorld, three mainstream GUI benchmarks spanning mobile and desktop platforms, EvoSkill-GUI consistently improves multiple base models without any training, with maximum gains of $+16.2\%$, $+6.0\%$, and $+10.5\%$ respectively, and evolved skill libraries continue to benefit related tasks rather than being rebuilt from scratch. Our code is available at \url{https://github.com/ZJU-REAL/EvoSkill-GUI}.

\end{abstract}

\input{section/intro}

\input{section/related_work}

\input{section/method}

\input{section/experiment}

\input{section/conclusion}

\clearpage
\section*{Limitations}
\label{sec:limitations}

EvoSkill-GUI relies on the backbone model to interpret screenshots, accessibility trees, and execution trajectories, and to produce structured revisions through tool calls. When the backbone's perception or trajectory diagnosis is unreliable, the resulting skill edit may not capture the true cause of failure and can in principle introduce regressions into a previously verified package. Although the tool-restricted interface and information-isolated critique reduce this risk, our framework does not currently include a formal verifier that vetoes harmful skill edits. Our evaluation focuses on three representative GUI benchmarks, but real-world deployment involves more frequent interface updates, personalized configurations, network-dependent latencies, and privacy-sensitive states than current benchmarks reflect. The reuse rate observed on AndroidWorld is achieved on parameterized variants of the same task family, and transfer to genuinely unseen application categories may degrade more gracefully than this number suggests. Finally, the skill-evolution loop incurs additional inference cost per failed rollout, namely one critic call plus one revision call; this overhead is negligible compared with retraining or fine-tuning the backbone model, but in latency-critical deployment it should be amortized across reuse.

\section*{Ethics Statement}
\label{sec:ethics}

EvoSkill-GUI is designed for benchmark GUI task execution and does not collect personal user data in our experiments. However, GUI agents that operate real applications can affect user accounts, files, settings, or third-party services if deployed without proper safeguards, so practical deployment should include explicit user authorization, action confirmation for sensitive operations, sandboxed execution when feasible, and audit logs of agent behavior. Skill packages produced by EvoSkill-GUI also warrant review before reuse in high-stakes or privacy-sensitive environments, since an incorrectly revised skill may silently repeat an unintended action across many future invocations; skill-level versioning, human-in-the-loop verification at deployment, and explicit fail-safe rules are natural next steps for responsible adoption of self-evolving GUI agents. Finally, we used large language models solely to assist with writing polish and minor code suggestions during implementation; all research ideas, framework design, experimental protocols, analysis, and conclusions, including the skill package design, the reflect-revise-reuse loop, the information-isolation protocol, and the retrieval mechanism, are the authors' own.

\bibliography{custom}

\clearpage
\appendix
\section{Appendix}

\subsection{Complete Implementation Details}
\label{implement}

Table~\ref{tab:implementation_hparams} reports the implementation details used in our experiments. 
For API-based models, we use the official provider endpoints with deterministic decoding. 
For locally deployed open-weight models, we use vLLM on one RTX PRO 6000 GPU 96GB server. 
We use public benchmarks and open-source tools only for research evaluation, following their released licenses and terms of use; API-based models are accessed through official provider endpoints under the corresponding provider terms, and our released code and prompts will include license information.
Unless otherwise specified, all benchmarks use the same structured skill-package format, retrieval strategy, and evaluation protocol.

\paragraph{Model serving.}
For GUI-Owl-1.5-8B-Instruct, Qwen3-VL-8B-Instruct and MAI-UI-8B, we serve the models with vLLM using tensor parallelism of 4, bfloat16 precision, maximum model length of 65,536, GPU memory utilization of 0.75, up to 20 concurrent sequences, maximum batched tokens of 65,536, and at most 5 images per prompt.

\paragraph{Evaluation.}
Each rollout is allowed up to 50 interaction steps and we set the per-step wait time to 3 seconds. 
The self-evolution loop uses at most three execution iterations, corresponding to the initial rollout plus at most two post-failure revision rounds.

\paragraph{Environment.}
We basically follow the default settings during experiments in all three benchmarks. For example, in MobileWorld experiments, we run 20 Docker containers in parallel with a 20-second launch interval, and set the maximum interaction rounds to 50 per task. For Qwen models, we use the original screenshot resolution $(1080 \times 2400)$ without resizing; for Claude models, we resize images to a maximum dimension of 1280 while preserving the aspect ratio (adaptive resize for Claude opus-4) or to a fixed size of $(1280, 720)$ for Claude sonnet-4. Action coordinates are normalized to a $[0, 1000]$ scale.

\paragraph{Excluded tasks.}
For the stability of our evaluation, we exclude 12 Google-proxy tasks from MobileWorld.

\subsection{Held-out skill transfer and retrieval quality}
\label{app:heldout_reuse}

\paragraph{Evolved skills transfer to held-out tasks.}
To evaluate whether the skill library benefits tasks beyond those used for its construction, we split MobileWorld into 87 library-building tasks and 30 held-out evaluation tasks.
On the held-out split, we compare a pass@3 baseline without skills against EvoSkill-GUI using only the library constructed from the 87 source tasks.
As shown in Table~\ref{tab:heldout_reuse}, EvoSkill-GUI solves 22/30 tasks ($73.3\%$), compared with 12/30 ($40.0\%$) for the pass@3 baseline.

\begin{table}[h]
\centering
\small
\begin{tabular}{lccc}
\toprule
\textbf{Retrieval result} & \textbf{Tasks} &
\textbf{Ours S/F} & \textbf{pass@3 S/F} \\
\midrule
Retrieved existing skill & 16 & 12/4 & 7/9 \\
No retrieval & 14 & 10/4 & 5/9 \\
\midrule
\textbf{Total} & \textbf{30} & \textbf{22/8} & \textbf{12/18} \\
\bottomrule
\end{tabular}
\caption{Held-out MobileWorld results. S/F denotes successful/failed tasks.}
\label{tab:heldout_reuse}
\end{table}

EvoSkill-GUI achieves ten additional successes overall.
Five come from tasks retrieving an existing skill, where EvoSkill-GUI solves 12/16 tasks compared with 7/16 for pass@3.
The remaining five come from tasks for which no skill is retrieved and a new package is generated.
This breakdown shows that the gain is not explained solely by generating a new skill for every held-out task; retrieved skills provide measurable transfer benefits.

\paragraph{Metadata-based retrieval remains precise on held-out tasks.}
Among the 16 tasks that trigger retrieval, 15 retrieve an appropriate skill ($93.8\%$), while only one produces a mismatch ($6.3\%$).
The mismatch occurs when \texttt{MattermostVisualInstructionResponseTask} retrieves an unrelated skill.
This result complements the metadata ablation in Table~\ref{tab:metadata_retrieval_ablation}, showing that structured metadata maintains high retrieval precision on held-out tasks.
When a mismatch occurs, the reflect--revise--reuse loop can revise the retrieved package from trajectory feedback, although such mismatches remain a potential source of negative transfer.

\subsection{Same-backbone self-diagnosis}
\label{app:self_diagnosis}

\paragraph{Same-backbone critique preserves self-contained evolution.}
EvoSkill-GUI deliberately uses the same backbone for execution and critique.
Introducing a stronger external critic would confound self-evolution with supervision from a more capable model.
The same-backbone design requires no additional model deployment or cross-model orchestration, although critic calls still incur the inference cost reported in Table~\ref{tab:token_cost}.

\paragraph{Information isolation separates critique from execution.}
Although the executor and critic share parameters, they operate in separate sessions with different information sets.
Under Eqs.~\ref{eq:isolation}, the critic receives only the instruction and execution trajectory, without access to the skill package, executor CoT, or ground truth.
It must therefore diagnose failures from observable trajectory evidence rather than inherit or rationalize the executor's original plan.

\paragraph{Same-backbone diagnosis is meaningful but imperfect.}
We manually annotate 143 critic judgments across 56 failed tasks.
As shown in Table~\ref{tab:critic_audit}, 115 judgments are correct, corresponding to an accuracy of $80.4\%$, while 28 are incorrect ($19.6\%$).

\begin{table}[h]
\centering
\small
\begin{tabular}{lcc}
\toprule
\textbf{Judgment} & \textbf{Count} & \textbf{Rate (\%)} \\
\midrule
Correct diagnosis & 115/143 & 80.4 \\
Incorrect diagnosis & 28/143 & 19.6 \\
\bottomrule
\end{tabular}
\caption{Manual audit of same-backbone critic judgments across 56 failed tasks.}
\label{tab:critic_audit}
\end{table}

The most consequential error is critic over-optimism: in at least four cases, the critic incorrectly judges a failed task as successful and terminates evolution prematurely.
These results show that the same backbone provides useful, but not infallible, diagnostic signals under information isolation.

\subsection{Analysis of negative OSWorld results}
\label{app:negative_osworld}

Table~\ref{tab:osworld} contains two negative domain-level entries, on OS for GUI-Owl-1.5-8B and Thunderbird for Qwen3-VL-8B-Instruct.
We identify two plausible factors behind these rare regressions.

\paragraph{Rare negative cases are sensitive to run-level variation.}
Across an evaluation spanning 369 tasks, only 3 tasks in two domain/model entries show negative changes, providing too few cases for a reliable general conclusion about negative transfer.
Moreover, aggregate scores on small domain splits can be sensitive to individual task outcomes.
For GUI-Owl-1.5-8B on the OS domain, two independent reruns both achieve $79.2\%$, whereas the submitted negative entry comes from a run in which EvoSkill-GUI fails one more task than the baseline.
This variation suggests that small domain-level differences should be interpreted cautiously.

\paragraph{Skill guidance can overcomplicate direct operations.}
EvoSkill-GUI is most beneficial for long workflows with stable interface structure and reusable operation patterns.
For tasks that admit a direct solution, however, imperfect skill guidance may introduce unnecessary intermediate steps or encourage over-reliance on an incorrect procedure.
This mismatch provides a plausible explanation for occasional negative transfer on tasks requiring short and precise execution.

\subsection{Token cost under repeated sampling.}

Table~\ref{tab:token_cost} reports token usage under a comparable test-time budget for pass@1 and pass@3 evaluation on MobileWorld.
The pass@3 baseline uses three independent rollouts, whereas EvoSkill-GUI performs three evolution rounds with skill retrieval, in-rollout revision, trajectory reflection, and skill-package updates.
EvoSkill-GUI consumes 94.75M tokens in total (0.810M per task and 0.439M per round), which is within the budget range of the pass@3 baseline (103.00M total and 0.880M per task).
The previously reported 168M tokens resulted from an accounting error in which intermediate cumulative snapshots were added to the final cumulative record, thereby double-counting tokens from earlier evolution rounds.
We correct this accounting error in the revised table.

\input{table/hyperpara}

\input{table/token_cost}

\paragraph{Accessibility trees add moderate runtime overhead but little token usage overhead.}
Table~\ref{tab:a11y_cost} further compares executions with and without accessibility-tree inputs on representative tasks. 
The token usage remains largely comparable across the two settings. 
By contrast, execution time is consistently higher with accessibility trees, mainly due to the additional collection and processing of structured UI information. 
This suggests that the main cost of accessibility-tree inputs comes from runtime overhead rather than prompt length, while their benefit is to provide more reliable grounding signals for GUI interaction.
\input{table/a11y_token_time}

\subsection{Case Studies}
\label{app:case_studies}

We provide two qualitative case studies to illustrate how EvoSkill-GUI turns critic diagnoses into concrete skill-package revisions. 
The examples cover two common failure patterns in long-horizon GUI tasks: missing verification checkpoints and incomplete traversal of scrollable content. 
In both cases, the agent initially fails despite following a plausible high-level plan; the isolated critic then diagnoses the failure from the executed trajectory, and EvoSkill-GUI writes the resulting correction back into the structured skill package. 
The later successful rollouts show how reflection-driven revision improves subsequent execution. 
The compact diagnosis-to-edit paths are visualized in Section~\ref{app:case_visualization}.

\subsubsection{Case 1: Verification Checkpoint Insertion}
\label{app:case_email}

\begin{figure*}[t]
    \centering
    \begin{subfigure}[t]{0.48\textwidth}
        \centering
        \includegraphics[width=\linewidth]{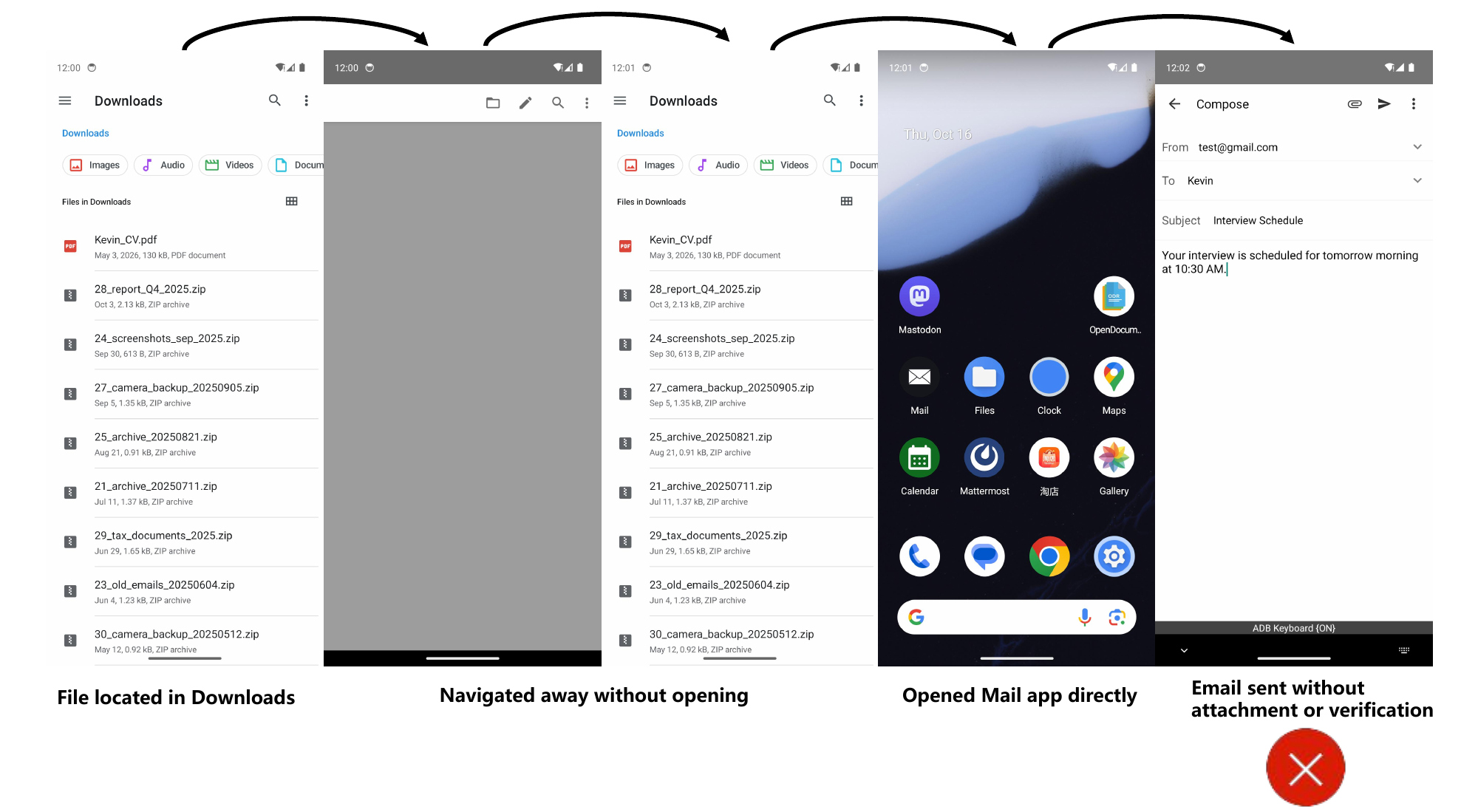}
        \caption{Failed rollout.}
        \label{fig:case1_fail}
    \end{subfigure}
    \hfill
    \begin{subfigure}[t]{0.48\textwidth}
        \centering
        \includegraphics[width=\linewidth]{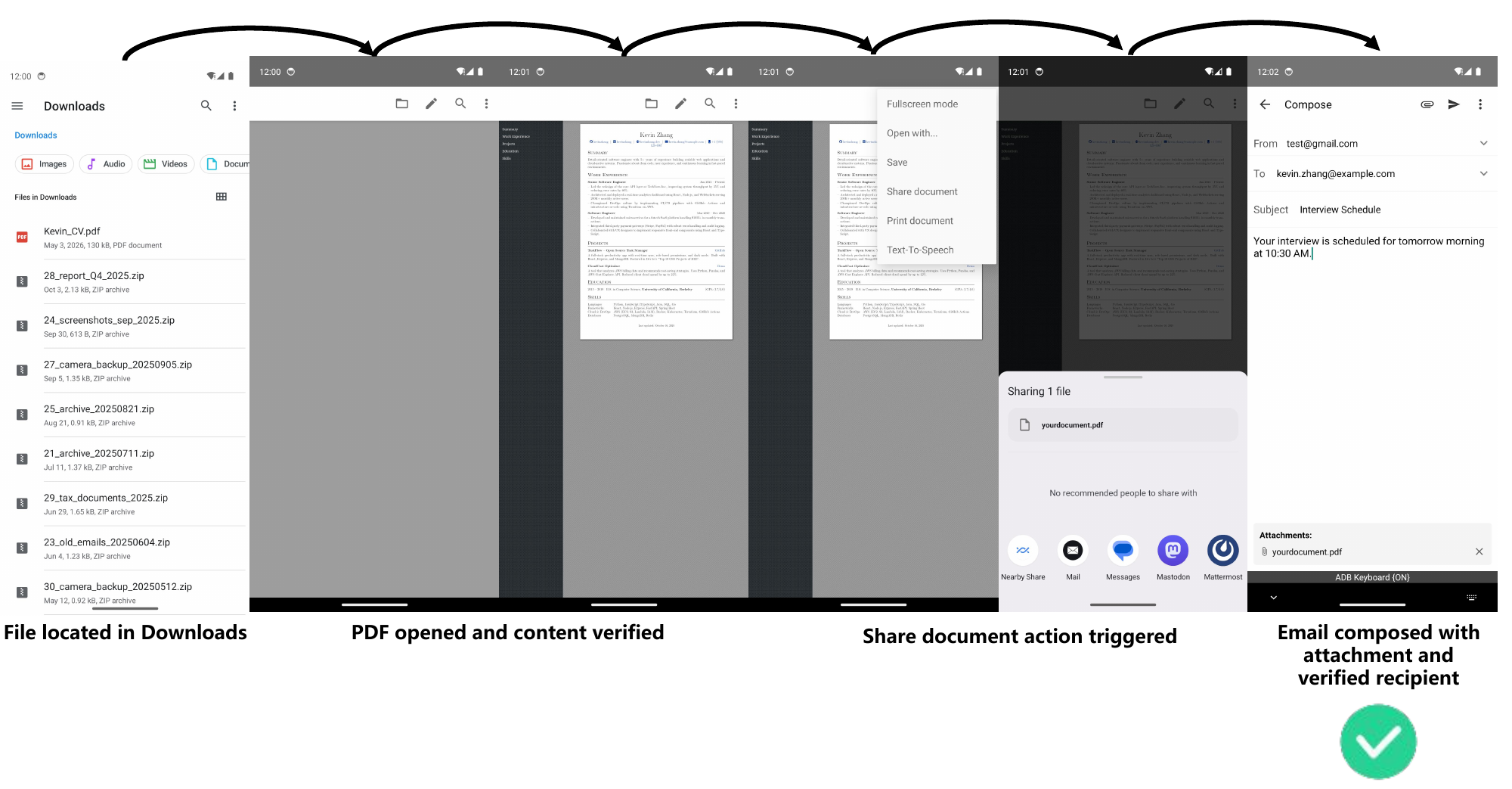}
        \caption{Successful rollout by \textbf{EvoSkill-GUI}.}
        \label{fig:case1_succ}
    \end{subfigure}
    \caption{Case study on \texttt{SendInterviewEmailTask}. \textbf{EvoSkill-GUI} revises the skill from a locate-only workflow into a locate--verify--attach workflow.}
    \label{fig:case1}
\end{figure*}

The first case studies \texttt{SendInterviewEmailTask}, where the agent must locate \texttt{Kevin\_CV.pdf}, verify that it is Kevin's resume, and send an interview email with the resume attached. 
As shown in Figure~\ref{fig:case1_fail}, the initial rollout fails because the agent treats locating the filename as sufficient evidence. 
It finds \texttt{Kevin\_CV.pdf} in the Downloads folder, but then leaves the file browser without opening the document, verifying its contents, extracting the recipient information, or attaching the resume to the email. 
This failure reveals a missing verification checkpoint rather than a simple grounding error: the plan contains the right target file, but lacks a required intermediate action before email composition.

The critic diagnoses this trajectory as a verification failure: the agent treats finding the filename as equivalent to confirming the document, and proceeds to email composition without opening the resume or checking whether the attachment and recipient are valid. 
This diagnosis is important because it identifies the missing operation at the procedural level rather than merely reporting that the final email was wrong.

Based on this critic diagnosis, EvoSkill-GUI revises the skill package by making verification and attachment explicit. 
The updated \texttt{plan.md} adds steps to open the PDF, confirm that the document is Kevin's resume, use the share sheet to attach the file to an email, and verify that the attachment is visible before sending. 
The same diagnosis also leads to new recovery rules for missing attachments and unresolved recipients, as well as backup localization instructions for validating the resume file and recipient field. 
With these diagnosis-driven edits, the second rollout succeeds, as shown in Figure~\ref{fig:case1_succ}. 
This case demonstrates that the critic can convert a failed trajectory into targeted procedural checkpoints that prevent premature task progression.

\begin{table}[h]
\centering

{\small
\setlength{\tabcolsep}{5pt}
\begin{tabular}{c c c l}
\toprule
\textbf{Round} & \textbf{Score} & \textbf{Steps} & \textbf{Failure Mode} \\
\midrule
1 & 0.0 & 14 & Skipped file verification \\
2 & 1.0 & 14 & Success \\
\bottomrule
\end{tabular}
}

\caption{Evolution summary for \texttt{Send\-Interview\-Email\-Task}.}
\label{tab:case1_summary}

\end{table}

\subsubsection{Case 2: Completeness Enforcement for List Traversal}
\label{app:case_cart}

\begin{figure*}[t]
    \centering
    \begin{subfigure}[t]{0.48\textwidth}
        \centering
        \includegraphics[width=\linewidth]{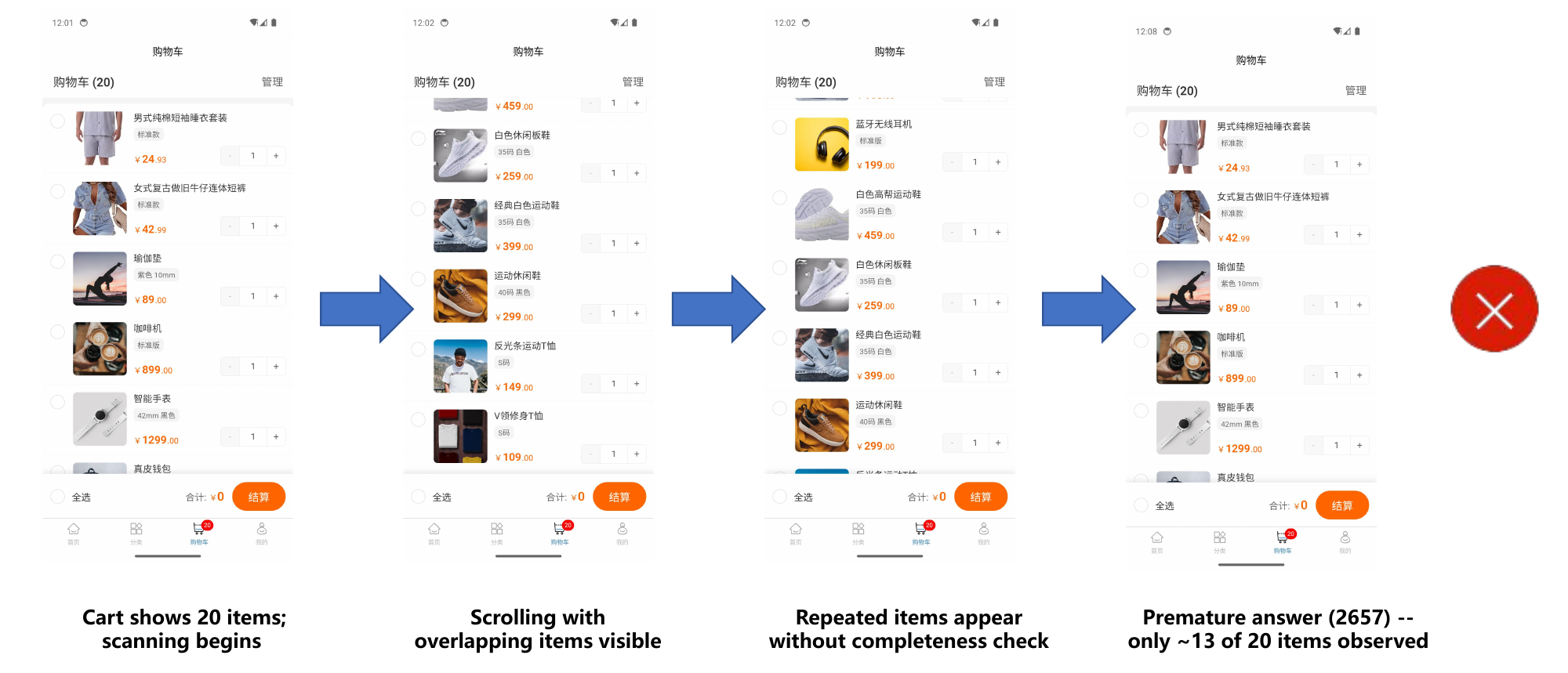}
        \caption{Failed rollout.}
        \label{fig:case2_fail}
    \end{subfigure}
    \hfill
    \begin{subfigure}[t]{0.48\textwidth}
        \centering
        \includegraphics[width=\linewidth]{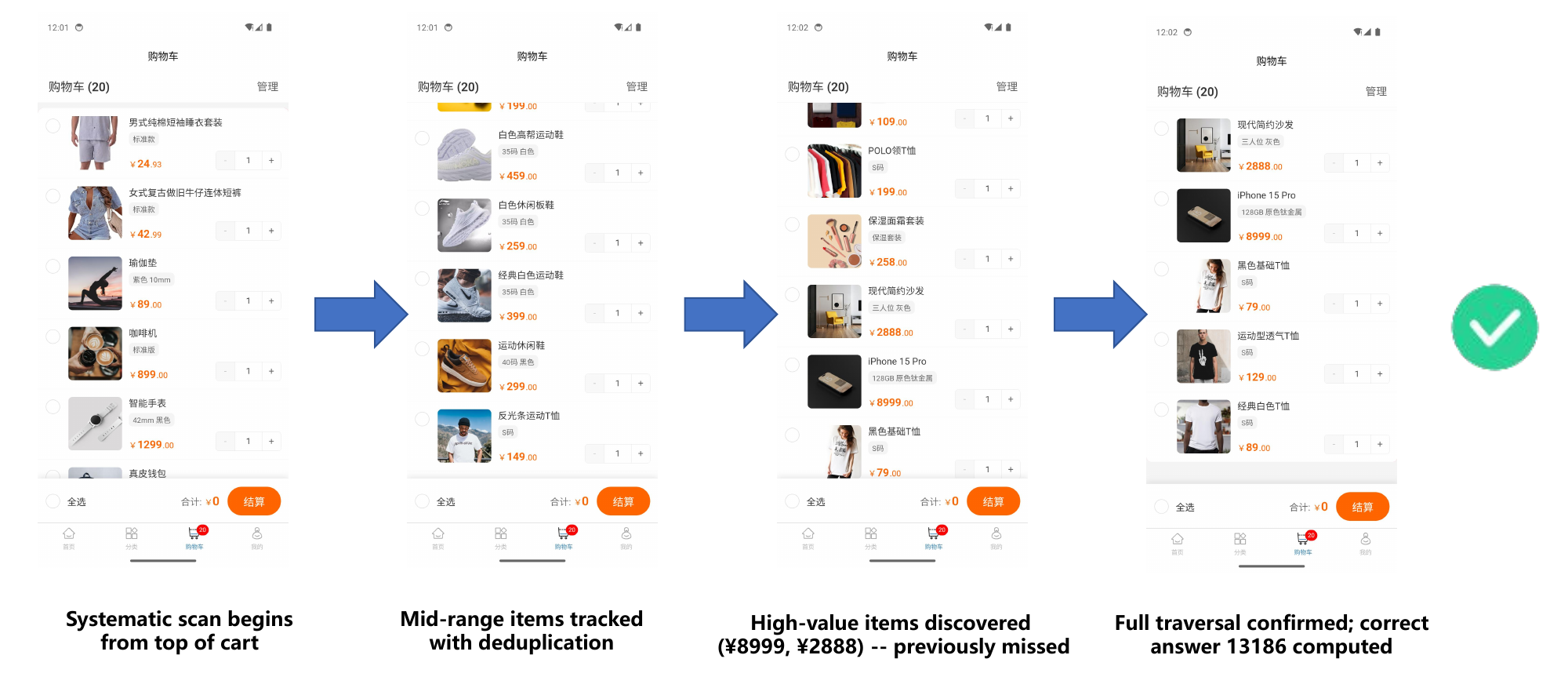}
        \caption{Successful rollout by \textbf{EvoSkill-GUI}.}
        \label{fig:case2_succ}
    \end{subfigure}
    \caption{Case study on \texttt{CheckCartPriceTask}. \textbf{EvoSkill-GUI} learns to verify complete traversal before reporting the top-3 item total.}
    \label{fig:case2}
\end{figure*}

The second case studies \texttt{CheckCartPriceTask}, where the agent must inspect a TaoDian cart with 20 items, identify the three most expensive items, and report their total price. 
The initial rollout fails because the agent scans only a partial view of the cart and answers based on three visible expensive items, producing 2657 instead of the correct total 13186. 
Although the agent scrolls through the list, the trajectory shows substantial overlap between views and no reliable evidence that all 20 items have been examined. 
This failure is therefore caused by incomplete traversal and missing termination criteria for a scrollable list.

The critic diagnoses the failure as incomplete traversal: the agent scrolls through the cart but never verifies that all 20 items have been examined before computing the top-3 total. 
This diagnosis separates the true failure cause from the surface behavior of ``scrolling many times'': the issue is not the absence of scrolling, but the absence of a completeness criterion.

Based on this diagnosis, EvoSkill-GUI first revises \texttt{plan.md} with systematic traversal logic: the agent must read the cart header, store the expected number of items, maintain a set of seen items, update a top-price list, and continue scrolling until both the end of the list and the expected item count are verified. 
It also updates \texttt{backup.md} with multiple end-of-list indicators, including visual markers, accessibility descriptions, repeated footer states, and cycle detection. 
However, the second rollout still repeats the same premature answer, showing that a single plan-level edit may be insufficient when the agent has already learned a misleading shortcut.

After the second failure, the critic again identifies the same root cause: the agent answers before proving that the cart traversal is complete. 
EvoSkill-GUI therefore appends a concrete negative example to \texttt{failure\_examples/}, recording that the answer 2657 was produced from an incomplete scan of roughly 13 unique items. 
This failure memory makes the critic's diagnosis persistent across later executions: the agent must not answer until it has counted all visible unique items or confirmed that no unseen items remain. 
In the third rollout, the agent successfully reports 13186 in 11 steps, as shown in Figure~\ref{fig:case2_succ}. 
This case shows that \texttt{failure\_examples/} complements plan revision by preserving critic-identified mistakes as explicit negative demonstrations.

\begin{table}[t]
\centering

{\small
\setlength{\tabcolsep}{5pt}
\begin{tabular}{c c c c c}
\toprule
\textbf{Round} & \textbf{Score} & \textbf{Steps} & \textbf{Answer} & \textbf{GT} \\
\midrule
1 & 0.0 & 25 & 2657 & 13186 \\
2 & 0.0 & 28 & 2657 & 13186 \\
3 & 1.0 & 11 & 13186 & 13186 \\
\bottomrule
\end{tabular}
}

\caption{Evolution summary for \texttt{CheckCartPrice\-Task}.}
\label{tab:case2_summary}

\end{table}

\subsubsection{Summary}
These cases show that EvoSkill-GUI improves skills by translating critic diagnoses into file-level updates of structured skill packages. 
In the email task, the critic identifies a missing verification step, which leads to new plan checkpoints and recovery rules for attachment and recipient validation. 
In the cart task, the critic identifies incomplete traversal, which first strengthens the traversal plan and later adds a concrete failure memory when the same mistake repeats. 
Together, the examples illustrate that EvoSkill-GUI does not merely retry failed tasks; it reflects on failed trajectories, identifies the procedural cause of failure, and revises the skill components that guide future execution.

\clearpage
\onecolumn
\subsubsection{Visualized Diagnosis-to-Edit Paths}
\label{app:case_visualization}

\subsubsection*{Case 1: SendInterviewEmailTask}
\begin{tracebox}{Round 1 Failure Trace}
\begin{itemize}
    \item Finds \texttt{Kevin\_CV.pdf} in Downloads.
    \item Leaves Files without opening the PDF.
    \item Sends email without content verification or attachment.
\end{itemize}
\end{tracebox}

\begin{criticbox}{Critic Diagnosis}
\textbf{Failure type:} missing verification checkpoint. \\
\textbf{Root cause:} the agent treats ``filename located'' as ``resume verified''. \\
\textbf{Required correction:} insert open--verify--attach checkpoints before email sending.
\end{criticbox}

\begin{skilldiffbox}{Skill Revision Based on Critic Diagnosis}
[plan.md]
+ Open Kevin_CV.pdf after locating it.
+ Verify the document is Kevin's resume.
+ Use Share -> Mail to attach the PDF.
+ Confirm attachment is visible before sending.
~ Recipient step: wait for a valid email autocomplete.

[recover.md]
+ If file is located but not opened: open and verify first.
+ If attachment is missing: return to Files and share again.
+ If recipient has no email: retype or extract from resume.

[backup.md]
+ Resume file: tap to open after locating.
+ To field: valid chip must include name and email.
\end{skilldiffbox}

\begin{successbox}{Round 2 Effect}
Locate-only workflow $\rightarrow$ locate--verify--attach workflow. 
The revised skill succeeds in 14 steps.
\end{successbox}
\clearpage

\subsubsection*{Case 2: CheckCartPriceTask}

\begin{tracebox}{Round 1 Failure Trace}
\begin{itemize}
    \item Scans only part of a 20-item cart.
    \item Computes top-3 total from visible items.
    \item Outputs 2657 instead of 13186.
\end{itemize}
\end{tracebox}

\begin{criticbox}{Critic Diagnosis after Round 1}
\textbf{Failure type:} incomplete traversal. \\
\textbf{Root cause:} scrolling is performed, but completeness is never verified. \\
\textbf{Required correction:} count unique items and stop only after end-of-list or total count is confirmed.
\end{criticbox}

\begin{skilldiffbox}{Skill Revision after Round 1}
[plan.md]
+ Read cart header and store TOTAL_ITEMS = 20.
+ Initialize SEEN_ITEMS and TOP_PRICES.
+ Add scroll-and-scan loop.
+ Stop only if end-of-list is verified OR len(SEEN_ITEMS) == TOTAL_ITEMS.
+ Answer only after traversal completeness is verified.

[backup.md]
+ Add visual end-of-list markers.
+ Add accessibility end-of-list cues.
+ Add OCR fallback for "no more".
+ Add cycle detection for repeated visible items.
\end{skilldiffbox}

\begin{tracebox}{Round 2 Failure Trace}
\begin{itemize}
    \item Repeats answer 2657.
    \item Scrolls extensively but does not count 20 unique items.
    \item Fails to prove that no unseen items remain.
\end{itemize}
\end{tracebox}

\begin{criticbox}{Critic Diagnosis after Round 2}
\textbf{Repeated root cause:} premature answer before complete traversal. \\
\textbf{Required correction:} preserve the failure as an explicit negative example.
\end{criticbox}

\begin{skilldiffbox}{Skill Revision after Round 2}
[failure_examples/failure_002.md]
+ Wrong answer: 2657.
+ Ground truth: 13186.
+ Diagnosis: answer was computed from visible items only.
+ Evidence: about 13 unique items observed, but cart has 20 items.
+ Rule: do not answer until all items are counted or end-of-list is verified.
\end{skilldiffbox}

\begin{successbox}{Round 3 Effect}
Partial scan shortcut $\rightarrow$ systematic traversal with failure memory. 
The revised skill reports 13186 in 11 steps.
\end{successbox}

\clearpage

\onecolumn
\subsection{Algorithm Pseudocode}
\begin{algorithm*}
\caption{EvoSkill-GUI}
\label{alg:evoskill}
\begin{algorithmic}[1]
\REQUIRE Instruction $I$, Environment $\mathcal{E}$, Skill Library $\mathcal{L}$, Policy $\pi_\theta$, Critic $\pi_\theta^J$, Max iterations $N$, Threshold $\theta_r$
\ENSURE Task completion trajectory $\tau$, Updated Skill Library $\mathcal{L}$

\STATE \textbf{Compute} $\text{score}(I, S)$ for all $S \in \mathcal{L}$ \hfill $\triangleright$ \textbf{Stage 1:} Retrieval (Eq.~\ref{eq:score})
\IF{$\max \text{score}(I, S) > \theta_r$}
    \STATE $S^{(0)} \gets \text{RetrieveTop}(I, \mathcal{L})$ \hfill $\triangleright$ Reuse verified skill package
\ELSE
    \STATE $S^{(0)} \gets \text{InitSkill}(D, A, P, B, C, F)$ \hfill $\triangleright$ Create new skill package (Eq.~\ref{eq:package})
\ENDIF

\STATE \hfill $\triangleright$ \textbf{Stage 2:} Self-evolution loop
\FOR{$i = 0$ \textbf{to} $N-1$}
    \STATE $\tau^{(i)}, S^{(i,\mathrm{end})} \gets \Phi(S^{(i)}, \mathcal{E})$ \hfill $\triangleright$ Rollout with instant revision (Eq.~\ref{eq:rollout})
    \IF{$\tau^{(i)}$ succeeds}
        \STATE $\mathcal{L} \gets \text{RegisterOrUpdate}(\mathcal{L}, S^{(i,\mathrm{end})})$ \hfill $\triangleright$ Register verified skill
        \STATE \textbf{return} $\tau^{(i)}, \mathcal{L}$
    \ELSE
        \STATE $c^{(i)} \sim \pi_{\theta}^{J}(\cdot \mid I, o_{1:T}^{(i)}, a_{1:T}^{(i)})$ \hfill $\triangleright$ Reflect: isolated diagnosis (Eq.~\ref{eq:critique})
        \STATE $S^{(i+1)} \sim \pi_{\theta}(\cdot \mid S^{(i,\mathrm{end})}, c^{(i)}, \tau^{(i)})$ \hfill $\triangleright$ Revise: edit skill package (Eq.~\ref{eq:revision})
        \STATE $S^{(i+1)}.F \gets S^{(i+1)}.F \cup \{(I, c^{(i)}, \tau^{(i)})\}$ \hfill $\triangleright$ Append failure case via \texttt{create\_failure}
    \ENDIF
\ENDFOR
\STATE $\mathcal{L} \gets \text{RegisterOrUpdate}(\mathcal{L}, S^{(N)})$ \hfill $\triangleright$ Save revised skill package
\STATE \textbf{return} $\tau^{(N-1)}, \mathcal{L}$

\end{algorithmic}
\end{algorithm*}

\clearpage

\subsection{Prompt Template}
\label{prompt}
\subsubsection{Executor System Prompts}
\tcbinputlisting{
    enhanced,
    breakable,
    listing only,
    title={EXECUTOR\_SKILL\_GENERATOR\_SYSTEM\_PROMPT\_JSON},
    colback=promptbg,
    colframe=promptframe,
    coltitle=white,
    colbacktitle=promptbar,
    fonttitle=\bfseries\ttfamily\small,
    arc=2mm,
    boxrule=0.5pt,
    left=1mm,
    right=1mm,
    top=1mm,
    bottom=1mm,
    listing file={appendix_prompts/executor_skill_generator.txt},
    listing options={style=promptstyle}
}

\tcbinputlisting{
    enhanced,
    breakable,
    listing only,
    title={EXECUTOR\_SKILL\_GENERATOR\_SYSTEM\_PROMPT\_XML},
    colback=promptbg,
    colframe=promptframe,
    coltitle=white,
    colbacktitle=promptbar,
    fonttitle=\bfseries\ttfamily\small,
    arc=2mm,
    boxrule=0.5pt,
    left=1mm,
    right=1mm,
    top=1mm,
    bottom=1mm,
    listing file={appendix_prompts/executor_skill_generator_xml.txt},
    listing options={style=promptstyle}
}

\subsubsection{Skill Generation Prompts}

\tcbinputlisting{
    enhanced,
    breakable,
    listing only,
    title={SEQUENTIAL\_SKILL\_GEN\_STEP\_1\_META\_PLAN\_PROMPT},
    colback=promptbg,
    colframe=promptframe,
    coltitle=white,
    colbacktitle=promptbar,
    fonttitle=\bfseries\ttfamily\small,
    arc=2mm,
    boxrule=0.5pt,
    left=1mm,
    right=1mm,
    top=1mm,
    bottom=1mm,
    listing file={appendix_prompts/step1.txt},
    listing options={style=promptstyle}
}
\tcbinputlisting{
    enhanced,
    breakable,
    listing only,
    title={SEQUENTIAL\_SKILL\_GEN\_STEP\_2\_BACKUP\_PROMPT},
    colback=promptbg,
    colframe=promptframe,
    coltitle=white,
    colbacktitle=promptbar,
    fonttitle=\bfseries\ttfamily\small,
    arc=2mm,
    boxrule=0.5pt,
    left=1mm,
    right=1mm,
    top=1mm,
    bottom=1mm,
    listing file={appendix_prompts/step2.txt},
    listing options={style=promptstyle}
}
\tcbinputlisting{
    enhanced,
    breakable,
    listing only,
    title={SEQUENTIAL\_SKILL\_GEN\_STEP\_3\_RECOVER\_PROMPT},
    colback=promptbg,
    colframe=promptframe,
    coltitle=white,
    colbacktitle=promptbar,
    fonttitle=\bfseries\ttfamily\small,
    arc=2mm,
    boxrule=0.5pt,
    left=1mm,
    right=1mm,
    top=1mm,
    bottom=1mm,
    listing file={appendix_prompts/step3.txt},
    listing options={style=promptstyle}
}

\subsubsection{Critic Reflection Prompts}
\tcbinputlisting{
    enhanced,
    breakable,
    listing only,
    title={CRITIC\_REFLECTION\_SYSTEM\_PROMPT},
    colback=promptbg,
    colframe=promptframe,
    coltitle=white,
    colbacktitle=promptbar,
    fonttitle=\bfseries\ttfamily\small,
    arc=2mm,
    boxrule=0.5pt,
    left=1mm,
    right=1mm,
    top=1mm,
    bottom=1mm,
    listing file={appendix_prompts/critic_sys.txt},
    listing options={style=promptstyle}
}
\tcbinputlisting{
    enhanced,
    breakable,
    listing only,
    title={CRITIC\_REFLECTION\_USER\_PROMPT},
    colback=promptbg,
    colframe=promptframe,
    coltitle=white,
    colbacktitle=promptbar,
    fonttitle=\bfseries\ttfamily\small,
    arc=2mm,
    boxrule=0.5pt,
    left=1mm,
    right=1mm,
    top=1mm,
    bottom=1mm,
    listing file={appendix_prompts/critic_user.txt},
    listing options={style=promptstyle}
}

\subsubsection{Skill Revision Prompts}
\tcbinputlisting{
    enhanced,
    breakable,
    listing only,
    title={SKILL\_REVISION\_SYSTEM\_PROMPT},
    colback=promptbg,
    colframe=promptframe,
    coltitle=white,
    colbacktitle=promptbar,
    fonttitle=\bfseries\ttfamily\small,
    arc=2mm,
    boxrule=0.5pt,
    left=1mm,
    right=1mm,
    top=1mm,
    bottom=1mm,
    listing file={appendix_prompts/skill_revise.txt},
    listing options={style=promptstyle}
}
\clearpage

\end{document}

%% file: section/intro.tex
\section{Introduction}
\label{sec:intro}

\begin{figure*}[!t]
    \centering
    
    \includegraphics[width=1.0\textwidth]{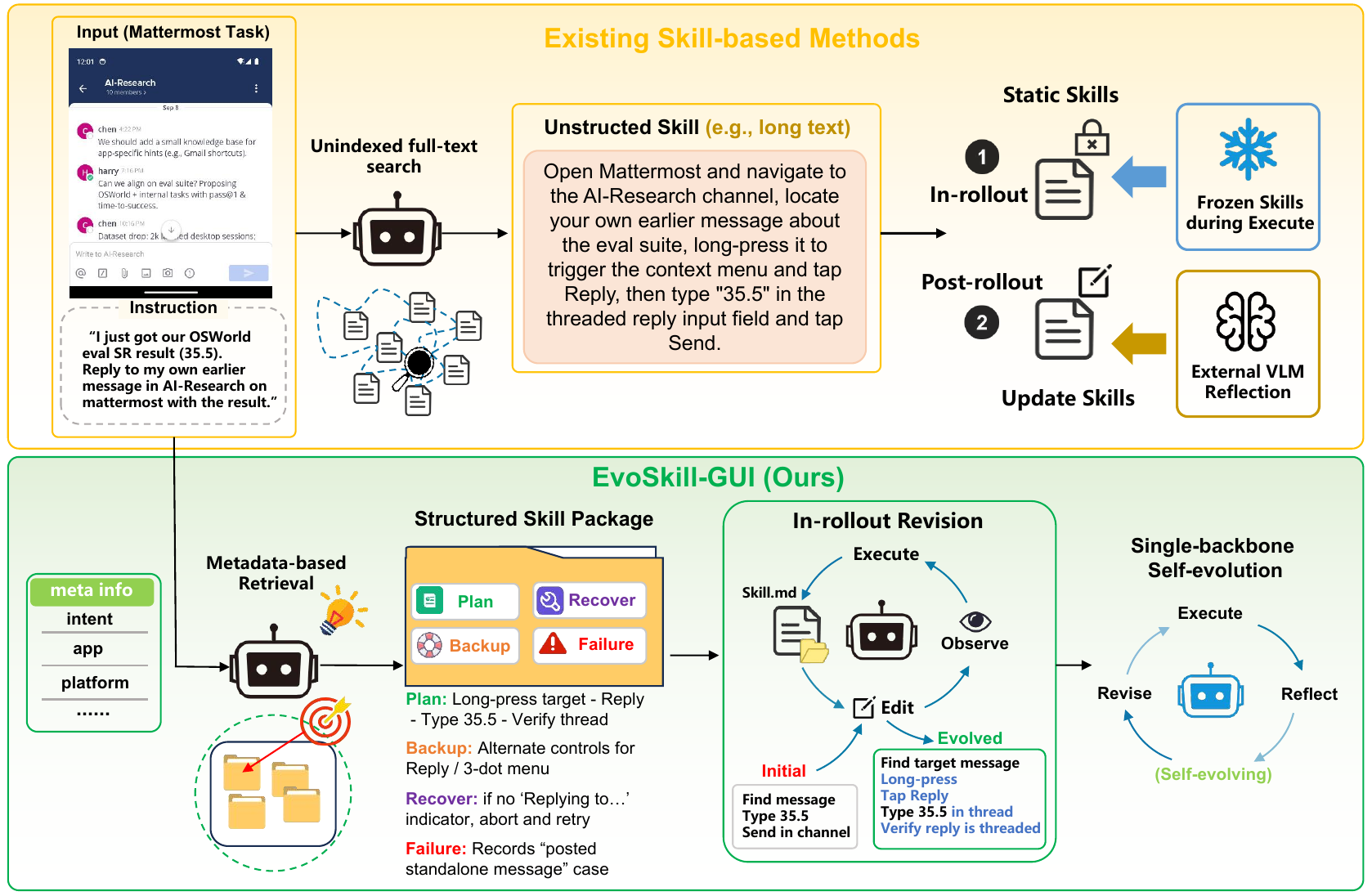}
    \vspace{-15pt}

    \caption{Comparison between existing skill-based methods and \textbf{EvoSkill-GUI}. 
    The upper panel illustrates the limitations of existing approaches, which rely on unstructured single-file skills, static skills during execution, external-VLM reflection, and unindexed full-text retrieval. 
    The lower panel shows how EvoSkill-GUI addresses these issues with structured editable skill packages, metadata-based retrieval, in-rollout revision, and single-backbone self-evolution, enabling skills to be adapted and reused from real execution feedback.}

    \vspace{-10pt}
    
    \label{fig:intro}
\end{figure*}

GUI agents~\cite{survey1,survey2,ye2025mobileagentv3} promise to automate long-horizon digital workflows on mobile and desktop applications. Unlike single-step execution tasks~\cite{screenspot,scrrenspotpro,lu2025uir1,tang2025guig2}, real GUI execution unfolds in non-stationary environments where pop-ups, delayed loads, and relocated widgets routinely invalidate plans fixed before execution~\cite{mobileuse,verigui2026}. Agent-skill frameworks~\cite{anthropic_skill,coevoskills,skillrl,survey_skills2026} provide a natural mechanism for reusing procedural knowledge across related tasks, yet existing skill designs are largely developed without targeting GUI execution dynamics, and treat skills as static artifacts produced before deployment rather than living procedural knowledge that improves through it.

Executing long-horizon GUI tasks under current skill-based paradigms encounters four barriers that existing designs~\cite{mmskills,cua_skill,webxskill} do not jointly address. \textbf{First, skills are often unstructured and hard to edit.} When procedural knowledge is stored as a single long document, planning steps, localization hints, and recovery rules are entangled, making targeted revision difficult. \textbf{Second, interfaces are non-stationary.} A skill valid at one interface state may fail when a pop-up appears, a widget shifts, or an accessibility tree becomes stale, and idempotent failure loops account for a substantial portion of timeouts in real GUI deployments~\cite{verigui2026,gui_reflection}. \textbf{Third, external reflection introduces extra cost and weak coupling.} Relying on an external reflection model adds latency and leaves reflection loosely coupled with skill execution and revision, making the system less fully self-evolving. \textbf{Fourth, procedural knowledge does not accumulate.} Lessons about unreliable selectors, missing contingencies, and required backtracks remain locked inside individual trajectories rather than becoming reusable assets for nearby tasks~\cite{memp2026,erl2026}. Together, these limitations indicate that GUI agents need skills that can not only be reused, but also revised from execution feedback.

Many GUI-agent failures carry actionable information about which part of a skill is wrong: grounding errors expose unreliable localization, unexpected pop-ups expose missing recovery branches, and redundant steps expose flawed plans.
Instead of discarding such failures or using them only for future training~\cite{seagent,evocua}, we ask \textbf{whether a deployed agent can immediately reflect on the failure and write the resulting procedural correction back into the skill itself}. Unlike concurrent self-evolving skill frameworks targeting code-centric environments~\cite{coevoskills} or structured agent memory updated through retrieval rather than skill-level revision~\cite{hymem2026,memp2026,memory_survey2026}, this yields a training-free formulation in which skills become structured, editable packages that accumulate task-specific experience over repeated use.

To realize this idea, we propose \textbf{EvoSkill-GUI}, a training-free self-evolving framework built around a \textbf{\emph{reflect-revise-reuse}} loop. Each skill is represented as a structured multi-file package containing retrieval metadata, executable plans, backup localization strategies, failure recovery rules, accessibility utilities, and failure examples. During rollout, the agent performs instant revisions when feedback contradicts the current plan. 
After a failed rollout, the same backbone model performs \textbf{single-backbone self-evolution}: it acts as an isolated critic to diagnose the trajectory under strict information isolation, and then as an executor to edit specific files of the package through a restricted tool interface so that revisions remain localized and auditable. Verified packages enter a library indexed by structured metadata, letting related tasks start from procedural knowledge that has already been validated.

We evaluate EvoSkill-GUI on MobileWorld, AndroidWorld, and OSWorld, three GUI benchmarks spanning mobile and desktop environments. Across both general-purpose and GUI-specialized base models, EvoSkill-GUI consistently improves task success rates without any training, with maximum gains of $+16.2\%$, $+6.0\%$, and $+10.5\%$ respectively. Ablations confirm the importance of structured packages, instant revision, and reflection-driven edits, and analyses indicate that evolved skill libraries continue to benefit related tasks rather than being rebuilt from scratch.

Our contributions are threefold:
\begin{itemize}
    \item We introduce EvoSkill-GUI, a training-free framework that enables GUI agents to construct, retrieve, execute, and revise self-evolving skill packages through information-isolated reflection and tool-restricted edits.
    \item We design a \textbf{GUI-oriented structured multi-file skill package} that separates retrieval metadata, executable plans, backup localization, recovery rules, accessibility utilities, and failure examples, enabling skills to be reused and updated at inference time.
    \item We demonstrate consistent improvements on MobileWorld, AndroidWorld, and OSWorld across multiple base models, with gains up to $+16.2\%$, $+6.0\%$, and $+10.5\%$ respectively.
\end{itemize}

%% file: section/related_work.tex
\section{Related Work}

\label{sec:related_work}
\subsection{Agent Skills}
Agent skills~\cite{anthropic_skill,surveyagentskills,survey_skills2026} externalize reusable procedural knowledge beyond model parameters, representing skills as structured artifacts with workflows, parameterized execution, and composition rules.
CUA-Skill~\cite{cua_skill} builds a structured desktop skill base, WebXSkill~\cite{webxskill} couples executable web programs with step-level guidance, and MMSkills~\cite{mmskills} extends skills with multimodal visual evidence.
XSkill~\cite{xskill} combines experiences and skills into a dual-stream continual learning framework updatable from past trajectories in a training-free manner.
Structured memory systems such as HyMem~\cite{hy_mem} and its GUI extension~\cite{hymem2026} organize reusable experience but focus on memory construction rather than skill-level revision.
At the learning level, SkillRL~\cite{skillrl}, D2Skill~\cite{d2skill}, Skill0~\cite{skill0}, and SDAR~\cite{lu2026sdar} explore skill distillation and internalization through reinforcement learning.
The concurrent CoEvoSkills~\cite{coevoskills} co-evolves multi-file skill packages with a surrogate verifier, but targets code-centric environments and skill construction rather than deployed GUI skill revision.
Overall, existing systems either construct skills offline, evolve them only during training, or keep them static after retrieval, leaving inference-time skill revision largely unexplored.

\subsection{Self-evolving Agents}
A separate line studies how agents improve from their own interaction experience.
In GUI settings, UI-Evol~\cite{ui_evol} refines external knowledge by retracing trajectories against reference knowledge, UI-Mem~\cite{ui_mem} maintains hierarchical experience memory over online reinforcement learning, and Mobile-Agent-E~\cite{mobile_agent_e} and SEAgent~\cite{seagent} evolve via persistent memory and reflective modules.
EvoCUA~\cite{evocua} couples synthetic experience generation with iterative policy evolution, turning failures into supervision through error analysis.
More recently, UI-Voyager~\cite{uivoyager2026} learns from failed mobile-GUI trajectories via group-relative self-distillation, MobileUse~\cite{mobileuse} adds hierarchical reflection at step and trajectory level, and GUI-Reflection~\cite{gui_reflection} embeds reflection supervision across pre-training, SFT, and online tuning.
Beyond GUI, Memp~\cite{memp2026} and Experiential Reflective Learning~\cite{erl2026} maintain procedural memory or reusable heuristics from past trajectories, alongside web and general-agent self-improvement~\cite{webevolver,uivoyager2026,hy_mem,memory_survey2026}.
These methods show that experience feedback supports continual improvement, but typically update memory, policy, or model weights rather than the deployed skill artifact itself.
In contrast, EvoSkill-GUI revises structured GUI skill packages from failure feedback, enabling self-evolution at the skill level without additional training.

\begin{figure*}[!t]
    \centering
    
    \includegraphics[width=1.0\textwidth]{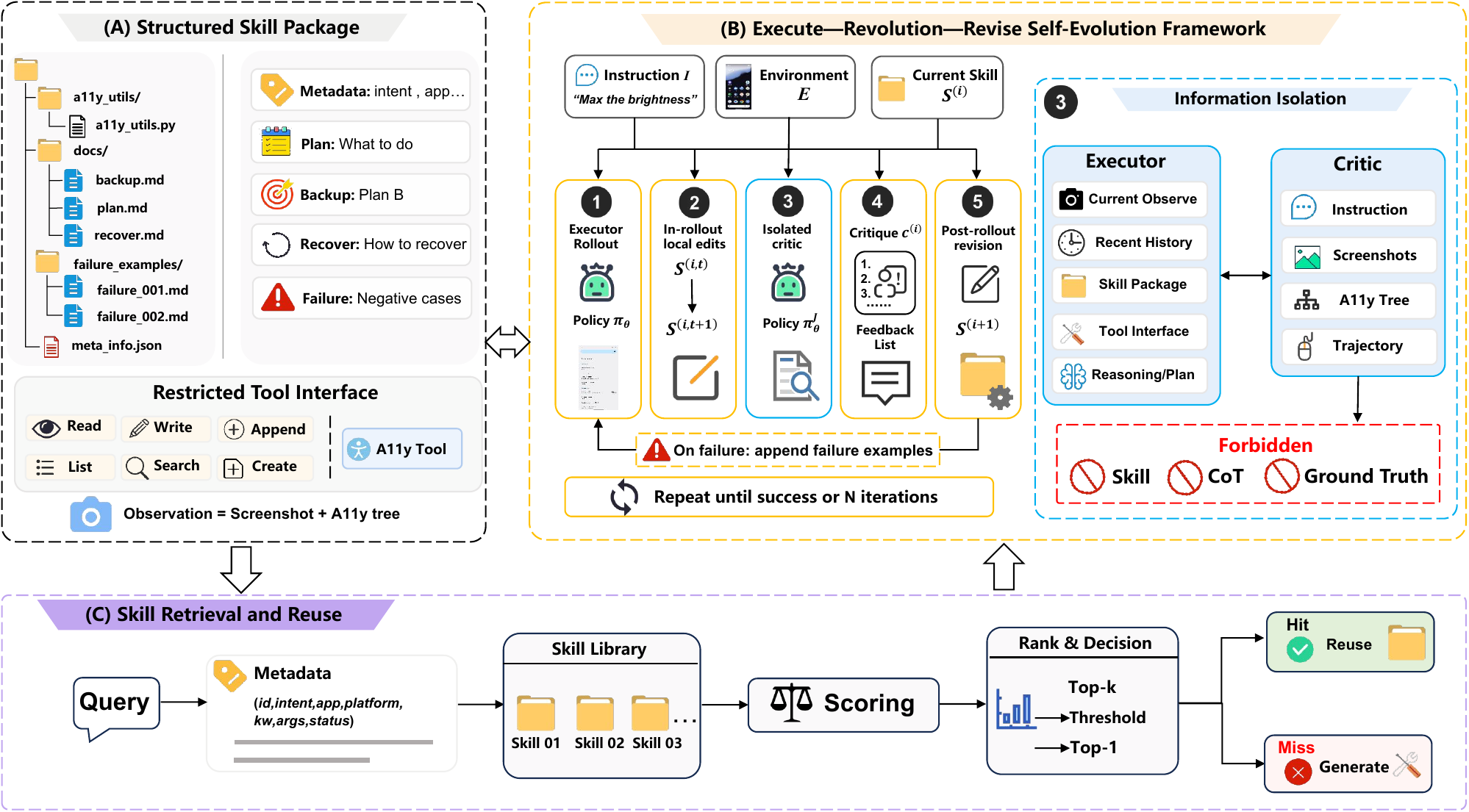}
    \caption{
Overview of \textbf{EvoSkill-GUI}.
\textbf{(A)} Each skill is a structured multi-file package containing accessibility utilities, planning, backup localization, recovery rules, failure cases, and metadata, accessed through a restricted tool interface.
\textbf{(B)} The execute--reflect--revise loop alternates rollouts with in-rollout edits, isolated post-rollout critique, and skill-file revisions, with strict information isolation between executor and critic.
\textbf{(C)} Evolved skills are indexed by metadata and reused via threshold-based scoring; misses trigger new skill generation.
}
    \vspace{-10pt}
    \label{fig:method}
\end{figure*}

%% file: section/method.tex
\section{Method}
\label{sec:method}

\subsection{Problem Formulation}
\label{sec:formulation}

Following a prior skill-evolution work's formulation~\cite{coevoskills}, we model GUI skill evolution as the problem of optimizing a reusable skill package under partial observability. 
A GUI task is modeled as a partially observable Markov decision process
$\mathcal{M}=\langle \mathcal{X}, \mathcal{A}, T, \mathcal{O}, \Omega, R\rangle$, where $\mathcal{X}$ denotes the underlying screen states, $\mathcal{A}$ denotes GUI actions and tool calls, $T$ is the transition function, and $R(x_T)\in\{0,1\}$ is the terminal task-completion reward. 
At each step, the agent observes $o_t=(\varphi_t,\zeta_t)$ consisting of a screenshot $\varphi_t$ and an accessibility tree $\zeta_t$, and acts based on the partial history $h_t=(o_1,a_1,\ldots,a_{t-1},o_t)$.

A skill package $S$ is a persistent procedural knowledge object that conditions the agent policy:
\begin{equation}
a_t \sim \pi_{\theta}(a_t \mid h_t,S).
\label{eq:policy}
\end{equation}
Given an instruction $I$, the expected task return of $S$ is
\begin{equation}
J(S,I)=
\mathbb{E}_{\tau \sim P(\tau \mid \pi_{\theta},S,\mathcal{M},I)}
\left[R(x_T)\right],
\label{eq:objective}
\end{equation}
where $\tau$ is the execution trajectory. 
For a task distribution $\mathcal{D}$, EvoSkill-GUI aims to obtain a reusable skill package that improves expected return across related tasks:
\begin{equation}
S^{*}=\arg\max_{S}\; \mathbb{E}_{I\sim \mathcal{D}}[J(S,I)].
\label{eq:argmax}
\end{equation}
The remaining question is how to represent $S$ and how to revise it without training or external supervision.

\subsection{Skill Package Representation}
\label{sec:skill}

EvoSkill-GUI represents each skill as a structured and editable package rather than a single free-form prompt. 
Formally, a skill package is
\begin{equation}
S=(D,A,P,B,C,F),
\label{eq:package}
\end{equation}
where $D$ is retrieval metadata, $A$ is accessibility tree related utilities, $P$ stores executable planning knowledge, $B$ stores backup localization and recognition strategies, $C$ stores failure-recovery rules, and $F=\{f^{(1)},\ldots,f^{(n)}\}$ stores failure cases. 
This decomposition aligns the package with three common sources of GUI-agent error: $P$ specifies what steps to perform, $B$ specifies how to identify and locate interface elements, and $C$ specifies how to recover when the expected state is violated. 
Compared with a monolithic skill file, this structure makes revisions more targeted: a grounding failure can update $B$, a missing contingency can update $C$, and a high-level procedural error can update $P$.

\paragraph{Tool interface.}To make the package editable during execution, EvoSkill-GUI exposes the skill directory through a restricted tool interface
\begin{equation}
\begin{aligned}
\mathcal{T}=\{&
\texttt{read},\texttt{write},\texttt{append}, \\
&\texttt{list},\texttt{search},\texttt{create\_failure}
\}.
\end{aligned}
\label{eq:tools}
\end{equation}
These tools allow the agent to inspect and revise the predefined skill files while preventing arbitrary modifications outside the package schema. 
In addition, a read-only accessibility tool $\psi$ returns the accessibility tree of the current window. 
Since $\psi$ only augments observation and does not modify $S$, EvoSkill-GUI separates interface understanding from skill revision at the tool level.

\subsection{Reflect-Revise Loop}
\label{sec:evo}

The loop alternates three phases: rollout with instant in-rollout revisions, isolated post-rollout reflection, and skill-file revision based on the critique.

\paragraph{Rollout and instant revision.}At iteration $i$, the executor runs the current skill package $S^{(i)}$ in the environment and obtains a trajectory
\begin{equation}
\tau^{(i)}=\Phi(S^{(i)},\mathcal{E}),
\label{eq:rollout}
\end{equation}
where $\Phi$ denotes rollout generation until success, failure, or the step limit. 
During the rollout, the executor may invoke tools in $\mathcal{T}$ to perform instant revisions, producing an intermediate package $S^{(i,\mathrm{end})}$. 
This allows the agent to correct local mismatches such as relocated widgets or unexpected pop-ups before they cascade into full task failure.

\paragraph{Reflection.}After each rollout, EvoSkill-GUI uses the same backbone model as an isolated critic $\pi_{\theta}^{J}$ to diagnose the trajectory:
\begin{equation}
c^{(i)} \sim \pi_{\theta}^{J}
(\cdot \mid I,o_{1:T}^{(i)},a_{1:T}^{(i)}).
\label{eq:critique}
\end{equation}
The critic outputs structured feedback, including failure-step localization, direct-cause analysis, and actionable revision suggestions. 
To prevent the critic from relying on privileged information unavailable to a deployed agent, we require the critic's information set $\mathcal{C}_{J}$ to be strictly contained in the executed trajectory:
\begin{subequations}
\label{eq:isolation}
\begin{align}
\mathcal{C}_{J} &\subseteq \{I\}\cup o_{1:T}\cup a_{1:T}, \\
\mathcal{C}_{J} &\cap \{S,\mathrm{CoT}_{\theta},\mathrm{GT}\}=\varnothing, \\
\mathcal{C}_{E} &\supseteq \mathcal{C}_{J}\cup\{S,\mathrm{CoT}_{\theta}\}.
\end{align}
\end{subequations}
The critic shares parameters with the executor but runs in a separate session, so improvements cannot be attributed to a stronger external supervisor.

\paragraph{Revision.}Given the rollout, the post-rollout skill snapshot, and the critic feedback, the executor revises the skill package:
\begin{equation}
S^{(i+1)} \sim \pi_{\theta}
(\cdot \mid S^{(i,\mathrm{end})},c^{(i)},\tau^{(i)}).
\label{eq:revision}
\end{equation}
If the rollout fails, EvoSkill-GUI appends a failure case $f^{(i)}=(I,c^{(i)},\tau^{(i)})$ to $F$ through \texttt{create\_failure}. 
The loop terminates when the task succeeds or when the maximum number of revision rounds is reached. 
In-rollout edits thus handle local execution mismatches, while post-rollout critique handles structural errors requiring trajectory-level diagnosis.

\subsection{Skill Retrieval and Reuse}
\label{sec:retrieval}

Evolved skills show much more value if they can be retrieved for later related tasks. 
EvoSkill-GUI therefore indexes each skill by structured metadata rather than by its full procedural body:

\begin{equation}
\begin{aligned}
D = (&\mathrm{id}, \mathrm{intent}, \mathrm{app}, \mathrm{platform}, \\
     &\mathrm{kw}, \mathrm{args}, \mathrm{hist}, \mathrm{status}).
\end{aligned}
\label{eq:metadata}
\end{equation}
where $\mathrm{intent}$ describes the skill goal, $\mathrm{app}$ and $\mathrm{platform}$ specify the execution context, $\mathrm{kw}$ stores retrieval keywords, $\mathrm{args}$ records reusable input slots, $\mathrm{hist}$ stores usage history, and $\mathrm{status}$ indicates whether the skill is verified. 
Indexing by metadata avoids matching against long procedural files whose surface tokens may overlap even across distinct tasks.

\paragraph{Retrieval score.}For a new instruction $q$, EvoSkill-GUI scores each candidate skill by combining lexical coverage, semantic tightness, and lightweight task constraints. 
With query tokens $Q$, skill tokens $T_S$, and overlap $O=Q\cap T_S$, the base score uses empirically chosen weights:
\begin{equation}
\mathrm{base}(q,S)=
0.6\frac{|O|}{|Q|}+0.4\frac{|O|}{|Q\cup T_S|}.
\label{eq:base}
\end{equation}
The final score adds bonuses for matched app and keyword fields and subtracts a divergence penalty for query-specific concept tokens that are missing from the candidate:
\begin{equation}
\small
\mathrm{score}(q,S)=
\mathrm{clip}(\mathrm{base}+b_{\mathrm{app}}+b_{\mathrm{kw}}-p_{\mathrm{div}},0,1).
\label{eq:score}
\end{equation}

\paragraph{Reuse.}At inference time, EvoSkill-GUI ranks the skill library by Eq.~\eqref{eq:score}. 
If the top candidate exceeds the retrieval threshold $\theta_r$, the agent reuses the corresponding skill package; otherwise, it creates a new package from scratch. 
A newly created package enters the library only after successful execution or after being revised through the self-evolution loop. 
Thus, future tasks can start from verified procedural knowledge instead of repeatedly rebuilding similar skills from zero.

%% file: section/experiment.tex
\section{Experiments}
\label{sec:experiment}

\subsection{Experimental Setup}
\paragraph{Benchmarks}
We evaluate EvoSkill-GUI on three interactive GUI benchmarks covering both mobile and desktop environments.
AndroidWorld~\cite{androidworld} provides a fully functional Android environment with 116 tasks across 20 real-world apps, enabling reproducible evaluation under dynamically parameterized instructions. 
MobileWorld~\cite{mobileworld} is a more challenging mobile-use benchmark designed to reflect realistic app usage through long-horizon, cross-app workflows. 
Since our focus is GUI interaction, we evaluate MobileWorld on GUI-only setting. 
OSWorld~\cite{osworld} provides a scalable real computer environment spanning Ubuntu, Windows, and macOS, supporting execution-based evaluation across arbitrary applications. 
Together, these benchmarks test whether EvoSkill-GUI can improve agents across platforms, applications, and task types.

\paragraph{Models}
We evaluate our method using two categories of base models: 
(1) general-purpose closed-weight models and open-weight models, including Claude-Sonnet-4.6~\cite{sonnet}, Qwen3.6-Plus, and Qwen3.6-35B-A3B~\cite{qwen3.6-35b-a3b}, which provide strong general multimodal reasoning; 
and (2) GUI-specialized open-weight models, including GUI-Owl-1.5-8B~\cite{gui_owl}, and MAI-UI-8B~\cite{mai_ui}, which are tailored for GUI understanding and grounding. 
This setup allows us to examine whether EvoSkill-GUI can consistently benefit both strong general agents and models already adapted for GUI interaction.
\paragraph{Implementations}
For closed-weight models and large-parameter open-weight models, we use the official APIs provided by their corresponding providers. 
For smaller open-weight models, we deploy them on one RTX PRO 6000 GPU 96GB using vLLM; the detailed deployment configuration is provided in the Appendix~\ref{implement}. 
Across all models and benchmarks, we keep the skill package format and retrieval strategy fixed. 
Each task is allowed up to 50 interaction steps, and after a failed attempt, the agent is allowed at most two rounds of skill revision. 
The prompts used in our experiments are also provided in the Appendix~\ref{prompt}.
\subsection{Main Results}

\input{table/osworld}

\input{table/androidworld}

\begin{figure}[!t]
    \centering
    \includegraphics[width=\columnwidth]{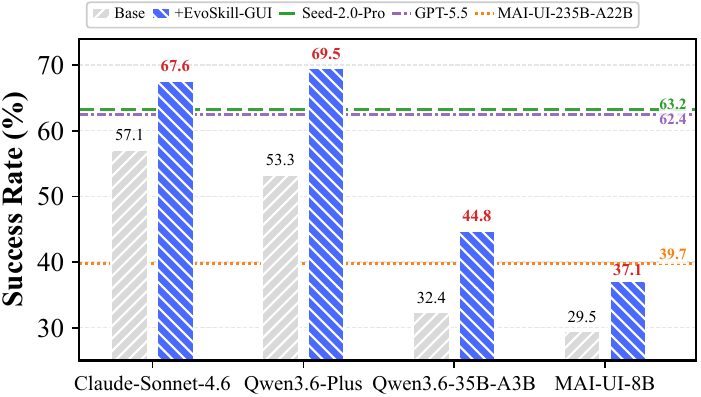}
    \vspace{-15pt}
    
    \caption{Performance comparison on MobileWorld under the GUI-only setting. The horizontal lines indicate representative reference baselines.}
    \vspace{-6pt}
    
    \label{fig:mobileworld}
\end{figure}

\paragraph{EvoSkill-GUI improves multiple base models on mobile GUI tasks.} 
On the GUI-only MobileWorld subset (Figure~\ref{fig:mobileworld}), EvoSkill-GUI improves every evaluated base model. 
For closed-weight models, it raises Claude-Sonnet-4.6 from $57.1\%$ to $67.6\%$ and Qwen3.6-Plus from $53.3\%$ to $69.5\%$. 
For open-weight models, it raises Qwen3.6-35B-A3B from $32.4\%$ to $44.8\%$ and MAI-UI-8B from $29.5\%$ to $37.1\%$. 
These consistent gains across model families and scales indicate that structured skill packages provide procedural guidance robust to the specific backbone.

Beyond MobileWorld, EvoSkill-GUI also improves AndroidWorld performance (Table~\ref{tab:androidworld_evoskill}), raising success rates by $+2.6\%$ under seed 30 and $+6.0\%$ under seed 42.

\paragraph{EvoSkill-GUI extends to desktop computer-use tasks.}
Table~\ref{tab:osworld} reports OSWorld performance on two base models. 
On GUI-Owl-1.5-8B, EvoSkill-GUI raises the overall success rate from $46.7\%$ to $54.8\%$ ($+8.1$), with large gains on applications such as Thunderbird ($+20.0$), VLC ($+42.7$), and VS Code ($+5.2$). 
On Qwen3-VL-8B-Instruct, the overall score improves from $23.8\%$ to $34.3\%$ ($+10.5$). 
These results show that the skill package design transfers beyond mobile environments to heterogeneous desktop workflows.
We discuss the two negative domain-level entries and their likely causes in Appendix~\ref{app:negative_osworld}.

\subsection{Ablation Study}
We conduct systematic ablation experiments on MobileWorld to validate the design of each component in EvoSkill-GUI.

\input{table/skill_organization}

\paragraph{Structured skill packages outperform monolithic skill files.}
Table~\ref{tab:structured_skill_ablation} compares the structured multi-file package with a single-file skill representation. 
The structured package improves success rate from $66.67\%$ to $69.52\%$ ($+2.85$). 
This supports our decomposition design: a monolithic file conflates planning, localization, and recovery, making targeted revision difficult, while the structured package allows each failure to update only the relevant component.

\input{table/iso_instant_merge}

\paragraph{Instant in-rollout revision prevents cascading failures.}
Table~\ref{tab:isolation_revision_ablation} shows that removing instant revision decreases MobileWorld success rate from $69.52\%$ to $62.86\%$ ($-6.66$). 
Relying on a fixed pre-generated plan throughout a long-horizon GUI task is unreliable, because intermediate states frequently diverge from the agent's original expectation by pop-ups, delayed loading, or changed widget positions.
Instant revision mitigates such mismatches before they cascade into full task failure.

\begin{figure*}[t]
    \centering
    \vspace{-8pt}
    
    \includegraphics[width=1\textwidth]{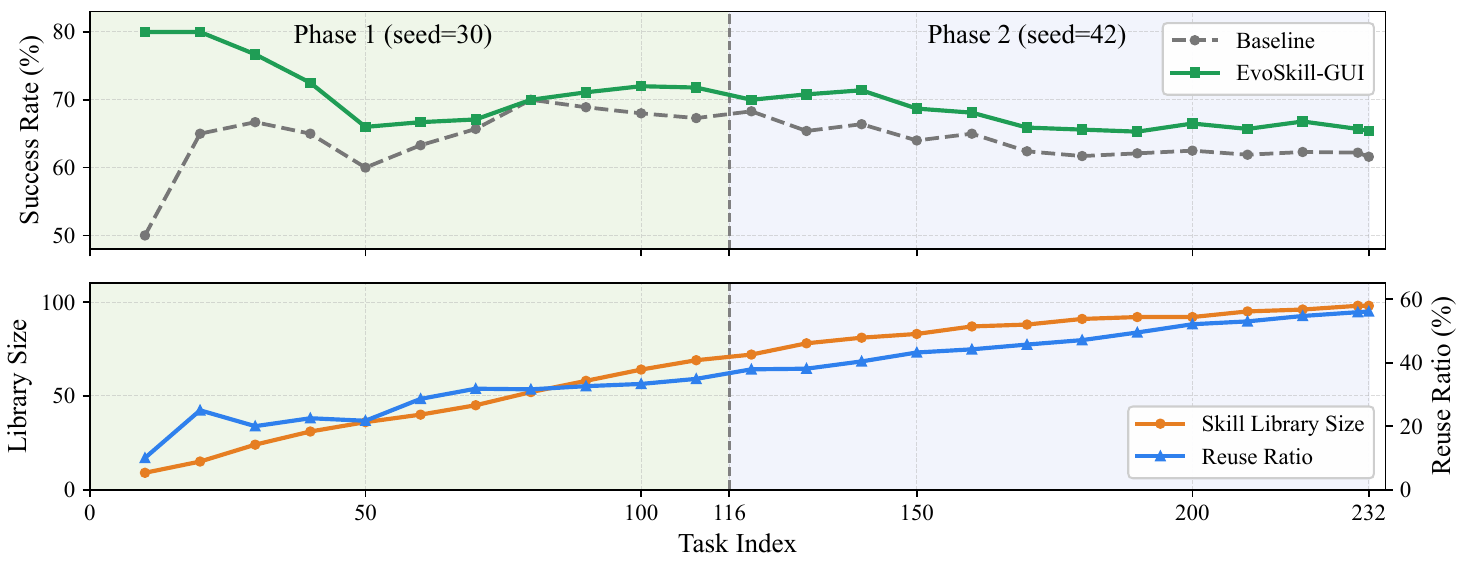}
    \vspace{-20pt}
    \caption{Skill-library growth on the 116+116 AndroidWorld task set. 
    EvoSkill-GUI maintains higher cumulative success rates than the baseline as the library grows and the reuse ratio increases.}
    \label{fig:skill_library_growth}
    \vspace{-10pt}
\end{figure*}

\begin{figure*}[b]
    \centering
    \begin{subfigure}[t]{0.49\textwidth}
        \centering
        \caption{Component ablation of the EvoSkill-GUI skill package.}
        \vspace{0pt}
        \includegraphics[width=\linewidth]{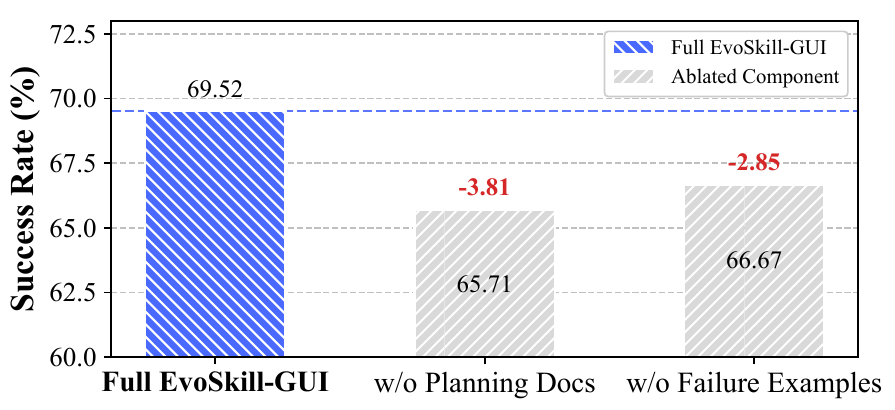}
        \vspace{-20pt}
        \label{fig:component}
    \end{subfigure}
    \hfill
    \begin{subfigure}[t]{0.5\textwidth}
        \centering
        \caption{Effect of accessibility tree inputs across different models.}
        \includegraphics[width=\linewidth]{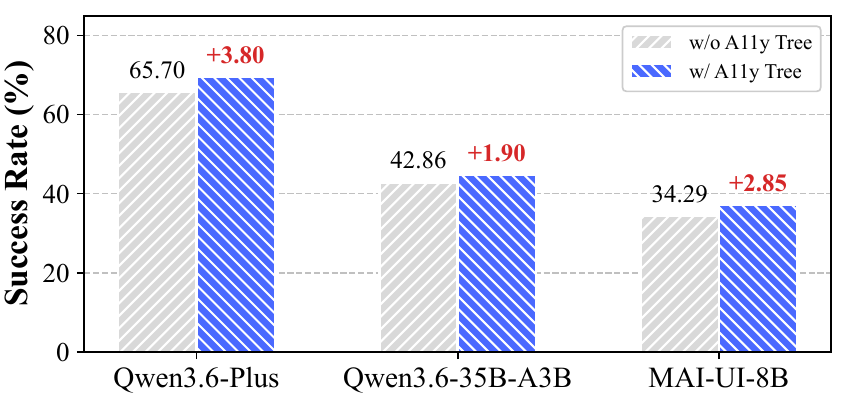}
        \vspace{-20pt}
        \label{fig:a11y}
    \end{subfigure}
    \vspace{-5pt}
    \caption{Ablation studies on skill-package components.}
    \vspace{-5pt}
    \label{fig:ablation_overview}
\end{figure*}

\paragraph{Metadata-based retrieval outperforms full-text matching.}
On twelve MobileWorld reuse cases shown in Table~\ref{tab:metadata_retrieval_ablation}, all completed successfully after retrieving the correct skill, metadata-based retrieval achieves an average score of $0.88$ compared with $0.41$ for full-text retrieval. 
Under retrieval threshold $\theta_r=0.6$, metadata-based retrieval recovers the correct skill in all $12$ cases, while full-text retrieval succeeds in only one. 
This supports structured metadata thus provides a strong retrieval signal that long full-text files, with overlapping surface tokens across distinct tasks, cannot reliably support.

\paragraph{Information isolation improves reflection quality.}

Table~\ref{tab:isolation_revision_ablation} shows that removing information isolation drops MobileWorld success rate from $69.52\%$ to $60.95\%$ ($-8.57$). 
By preventing the critic from relying on privileged context or inheriting the executor's mistaken plan, information isolation produces more evidence-grounded diagnoses and leads to more reliable skill revision.

\begin{figure}[t]
    \centering
    \vspace{-8pt}
    
    \includegraphics[width=0.48\textwidth]{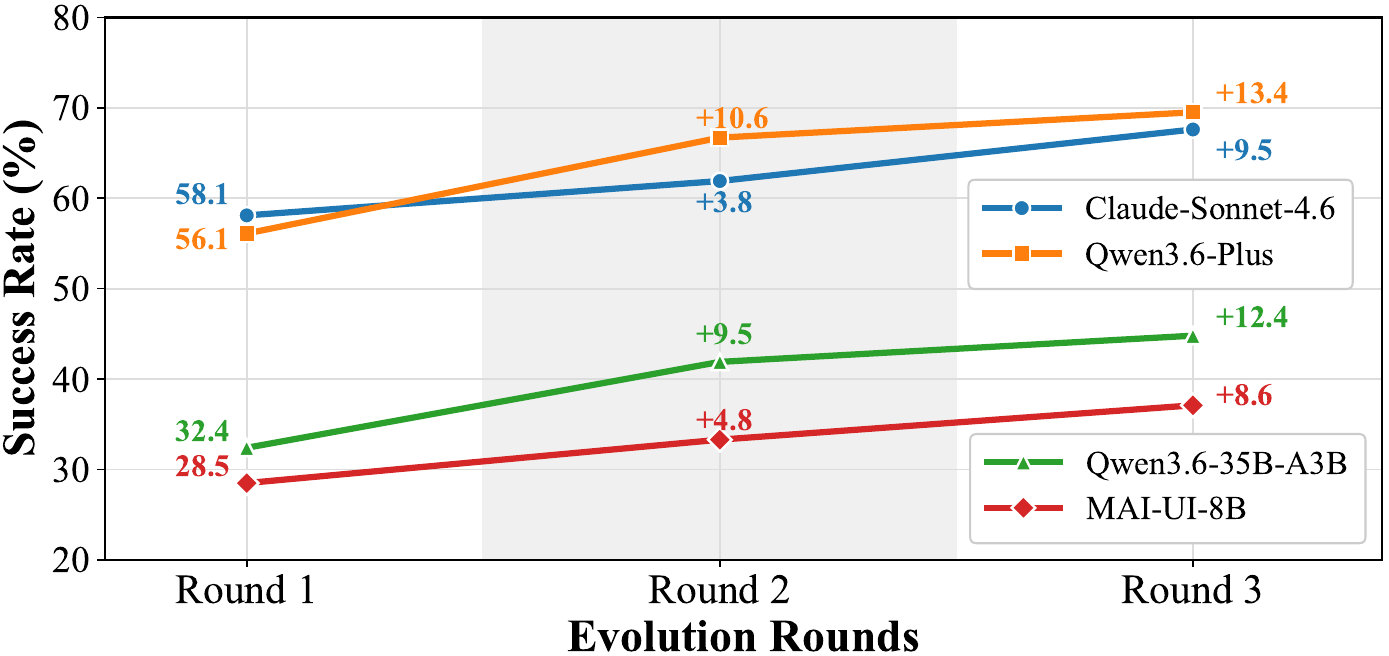}
    \vspace{-15pt}
    \caption{Three round evolution of multiple base models on MobileWorld.}
    \vspace{-8pt}
    \label{fig:evolution}
\end{figure}

\input{table/meta_info}

\paragraph{Each skill-package component contributes to performance.}
Figure~\ref{fig:component} ablates skill-package components. 
Removing planning documents reduces success rate from $69.52\%$ to $65.71\%$ ($-3.81$), the largest drop, indicating that explicit procedural plans are critical for maintaining a coherent action sequence across long horizons. 
Removing failure reflection also drops performance to $66.67\%$ ($-2.85$), showing that post-failure diagnosis provides useful corrective knowledge for subsequent executions.

\paragraph{Accessibility-tree utilities benefit grounding across models.}
Figure~\ref{fig:a11y} shows that adding accessibility-tree inputs improves Qwen3.6-Plus by $+3.80$, Qwen3.6-35B-A3B by $+1.90$, and MAI-UI-8B by $+2.85$. 
The consistent gains across backbone families suggest that accessibility-tree utilities act as a structured observation channel complementary to vision, supplying actionable grounding cues regardless of the model's GUI pretraining.

\paragraph{A moderate retrieval threshold balances skill reuse and selectivity.}
Table~\ref{tab:retrieval_threshold} analyzes the sensitivity of EvoSkill-GUI to the retrieval threshold $\theta_r$ on MobileWorld.
Lowering the threshold from $0.6$ to $0.4$ reduces the success rate from $69.5\%$ to $55.2\%$, suggesting that overly permissive retrieval can introduce weakly related skill packages.
Raising the threshold to $0.8$ also decreases the success rate to $66.7\%$, indicating that overly conservative retrieval may miss useful reusable skills.
The best performance at $\theta_r=0.6$ supports its use as a balanced default in our experiments.

\input{table/retrieval}

\subsection{Analysis}

\paragraph{Skill evolution provides cumulative gains and largely saturates after three rounds.}
Figure~\ref{fig:evolution} shows that EvoSkill-GUI consistently improves success rates across rounds on MobileWorld.
Claude-Sonnet-4.6 rises from $58.1\%$ in Round 1 to $67.6\%$ in Round 3.
The open-weight models show the same trend: Qwen3.6-35B-A3B improves from $32.4\%$ to $44.8\%$, and MAI-UI-8B from $28.5\%$ to $37.1\%$.
We further extend Qwen3.6-Plus to five rounds to examine saturation.
Its success rate rises from $56.2\%$ to $61.9\%$ ($+5.7$) and $68.6\%$ ($+6.7$) over the first three rounds, remains at $68.6\%$ in Round 4, and reaches $69.5\%$ in Round 5 ($+0.9$).
Thus, the first three rounds yield a $12.4$-point gain, while two additional rounds add only $0.9$ point, supporting three rounds as an effective performance--cost trade-off.

\input{table/reuse}

\input{table/retrieval}
\input{table/repair_analysis}

\paragraph{Skill evolution outperforms repeated sampling under a comparable budget.}
Table~\ref{tab:token_cost} compares EvoSkill-GUI with a Qwen3.6-Plus pass@3 baseline on MobileWorld.
While three independent baseline rollouts achieve $62.8\%$ using 103.00M tokens, three-round EvoSkill-GUI achieves $69.5\%$ using 94.75M tokens, improving accuracy by $6.7$ points with a slightly lower budget.
This comparison suggests that the gains arise from reflection-driven skill evolution rather than repeated sampling or increased token usage alone.

\paragraph{Evolved skill libraries remain reusable and effective on related tasks.}
Tables~\ref{tab:androidworld_evoskill} and~\ref{tab:failure_recovery} analyze whether EvoSkill-GUI accumulates reusable procedural knowledge across related AndroidWorld tasks. 
We build a 116$+$116 task set with seed-30 tasks as Phase 1 and related seed-42 variants as Phase 2. 
EvoSkill-GUI improves success rate from $68.1\%$ to $70.7\%$ in Phase 1, where $37.9\%$ of tasks already reuse matched skills, and from $55.2\%$ to $61.2\%$ in Phase 2, where all tasks reuse the evolved library. 
Table~\ref{tab:failure_recovery} further shows that these reused skills remain effective: EvoSkill-GUI recovers 27 out of 89 initially failed executions overall, including 19 out of 65 failures by reused skills. 
These results indicate that the skill library accumulates reusable procedural experience over time, rather than rebuilding one-off task solutions from scratch.

\paragraph{Skill libraries accumulate reusable experience over task streams.}
Figure~\ref{fig:skill_library_growth} shows EvoSkill-GUI on the sequential 116+116 AndroidWorld task set. 
In Phase 1, the library grows from 9 skills at task 10 to 69 skills at task 110, with reuse becoming more frequent as more matches become available. 
In Phase 2, the evolved library is applied to related seed-42 variants, further reaching 98 skills and a 56.1\% reuse ratio. 
The higher cumulative success rates across both phases suggest that EvoSkill-GUI accumulates reusable procedural experience rather than isolated task solutions. 
This motivates the following recovery analysis.

\paragraph{Different failure types require different skill edits.}
We next ask which failures are actually repairable by skill revision. 
Table~\ref{tab:repairability_analysis} shows that plan-level errors are the most repairable, with 16/23 recovered by revising \texttt{plan.md}; grounding errors are partially repairable through \texttt{backup.md} (6/12), while missing contingencies are harder to fix through \texttt{recover.md} (1/4). 
Overall, this supports the structured design of EvoSkill-GUI: separating planning, grounding, and recovery knowledge lets revision target the component most responsible for a failure.

%% file: table/osworld.tex
\begin{table*}[htbp]
    \centering
    
    \resizebox{\textwidth}{!}{
    \setlength{\tabcolsep}{3pt}
    \begin{tabular}{@{}l*{11}{c}@{}}
        \toprule
        \multirow{2}{*}{\textbf{Method}} & \multirow{2}{*}{\textbf{Chrome}} & \multirow{2}{*}{\textbf{GIMP}} & \multicolumn{3}{c}{\textbf{LibreOffice}} & \multirow{2}{*}{\textbf{Multi-Apps}} & \multirow{2}{*}{\textbf{OS}} & \multirow{2}{*}{\textbf{Thunderbird}} & \multirow{2}{*}{\textbf{VLC}} & \multirow{2}{*}{\textbf{VS Code}} & \multirow{2}{*}{\textbf{Overall}} \\
        \cmidrule(lr){4-6}
         & & & \textbf{Calc} & \textbf{Impress} & \textbf{Writer} & & & & & & \\
        \midrule

        \rowcolor{gray!10}
        \multicolumn{12}{l}{\textit{General Closed-Weights Models}} \\

        Claude-Sonnet-4.5~\cite{sonnet}
        & 56.4 & 57.7 & 66.0 & 57.5 & 65.2 & 47.0 & 70.8 & 66.7 & 52.9 & 69.6 & 58.1 \\

        Qwen3-VL-Flash~\cite{qwen3vl}
        & 56.4 & 42.3 & 23.0 & 51.0 & 56.5 & 22.0 & 54.2 & 66.7 & 34.4 & 69.6 & 41.6 \\

        \midrule

        \rowcolor{gray!10}
        \multicolumn{12}{l}{\textit{Specialized Open-Weights Models}} \\

        UI-TARS-72B-DPO~\cite{uitars}
        & 33.2 & 61.5 & 12.8 & 25.4 & 43.5 & 6.7 & 33.3 & 33.3 & 23.5 & 47.8 & 25.9 \\
        Computer-Use-Preview
        & 36.9 & 34.6 & 14.9 & 29.7 & 26.1 & 15.8 & 70.8 & 66.7 & 11.8 & 69.6 & 31.2 \\
        
        OpenCUA-32B~\cite{opencua}
        & 45.6 & 65.4 & 12.8 & 42.5 & 39.1 & 14.6 & 60.9 & 46.7 & 23.5 & 65.2 & 35.1 \\

        EvoCUA-8B~\cite{evocua}
        & 54.3 & 84.6 & 29.8 & 38.2 & 52.2 & 25.5 & 75.0 & 73.3 & 44.8 & 65.2 & 46.1 \\

        EvoCUA-32B 
        & 65.1 & 76.9 & 55.3 & 59.5 & 69.6 & 27.9 & 78.3 & 80.0 & 49.4 & 87.0 & 56.7 \\

        GUI-Owl-1.5-32B~\cite{gui_owl}
        & 66.6 & 76.9 & 48.9 & 63.8 & 60.9 & 27.3 & 76.2 & 86.7 & 37.4 & 87.0 & 55.4 \\

        \midrule
        \rowcolor{gray!10}
        \multicolumn{12}{l}{\textit{Skill-based Works}} \\
        MMSkills~\cite{mmskills}
        & 39.9 & 42.3 & 8.5 & 23.4 & 17.4 & 13.4 & 25.0 & 29.4 & & 47.8 & 25.4 \\
        \midrule
        
        \rowcolor{gray!10}
        \multicolumn{12}{l}{\textit{Our Method}} \\

        GUI-Owl-1.5-8B
        & 35.6 & 80.8 & 29.8 & 57.5 & 69.6 & 21.5 & 70.8 & 73.3 & 43.8 & 81.8 & 46.7 \\

        \textbf{+EvoSkill-GUI}
        & \textbf{41.3} & \textbf{88.5} & \textbf{37.8} & \textbf{65.8} & \textbf{69.6} & \textbf{30.5} & \textbf{66.7} & \textbf{93.3} & \textbf{86.5} & \textbf{87.0} & \textbf{54.8} \\

        \textcolor{darkgreen}{$\Delta$\textit{Improvement}}
        & \textcolor{darkgreen}{+5.7} & \textcolor{darkgreen}{+7.7} & \textcolor{darkgreen}{+8.0} & \textcolor{darkgreen}{+8.3} & \textcolor{darkgreen}{+0.0} & \textcolor{darkgreen}{+9.0} & \textcolor{darkgreen}{-4.1} & \textcolor{darkgreen}{+20.0} & \textcolor{darkgreen}{+42.7} & \textcolor{darkgreen}{+5.2} & \textcolor{darkgreen}{+8.1} \\

        \midrule

        Qwen3-VL-8B-Instruct 
        & 28.1 & 26.9 & 19.1 & 31.8 & 47.8 & 7.7 & 16.7 & 60.0 & 35.3 & 26.1 & 23.8 \\

        \textbf{+EvoSkill-GUI}
        & \textbf{43.4} & \textbf{46.2} & \textbf{29.8} & \textbf{34.4} & \textbf{52.2} & \textbf{15.2} & \textbf{41.7} & \textbf{46.7} & \textbf{50.6} & \textbf{47.8} & \textbf{34.3} \\

        \textcolor{darkgreen}{$\Delta$\textit{Improvement}}
        & \textcolor{darkgreen}{+15.3} & \textcolor{darkgreen}{+19.3} & \textcolor{darkgreen}{+10.7} & \textcolor{darkgreen}{+2.6} & \textcolor{darkgreen}{+4.4} & \textcolor{darkgreen}{+7.5} & \textcolor{darkgreen}{+25.0} & \textcolor{darkgreen}{-13.3} & \textcolor{darkgreen}{+15.3} & \textcolor{darkgreen}{+21.7} & \textcolor{darkgreen}{+10.5} \\

        \bottomrule
    \end{tabular}
    \vspace{-25pt}
    
    }

    \caption{
    Performance comparison on the OSWorld-Verified benchmark across two models:
    GUI-Owl-1.5-8b and Qwen3-VL-8B-Instruct.
    For a fair comparison, all results are evaluated under a maximum of 50 steps.
    }
    \vspace{-5pt}
    \label{tab:osworld}
\end{table*}

%% file: table/androidworld.tex
\begin{table}[t]
    \centering
    \small
    \setlength{\tabcolsep}{5pt}
    \begin{tabular}{l c c c c}
    \toprule
    \textbf{Phase} & \textbf{Reuse Rate} & \textbf{Base SR} & \textbf{Ours SR} & \textbf{$\Delta$} \\
    \midrule
    P1 (seed 30) & 37.9  & 68.1 & \textbf{70.7} & \textcolor{darkgreen}{+2.6} \\
    P2 (seed 42) & 100.0 & 55.2 & \textbf{61.2} & \textcolor{darkgreen}{+6.0} \\
    \bottomrule
    \end{tabular}
    \caption{Evaluation on the AndroidWorld task set. SR denotes success rate.}
    \vspace{-12pt}
    \label{tab:androidworld_evoskill}
    
\end{table}

%% file: table/skill_organization.tex
\begin{table}[htbp]
  \centering
  \small
  \setlength{\tabcolsep}{5pt}
  
  \begin{tabular}{lc}
    \toprule
    \textbf{Skill Organization} & \textbf{Success Rate (\%)} \\
    \midrule
    Single-file Skill & 66.67 \\
    \textbf{Skill Package (Ours)} & \textbf{69.52} \\
    \midrule
    \textcolor{darkgreen}{$\Delta$ \textit{Improvement}} & \textcolor{darkgreen}{+2.85} \\
    \bottomrule
    \vspace{-10pt}
  \end{tabular}
  \caption{Ablation of skill organization on MobileWorld.}
  \label{tab:structured_skill_ablation}
  \vspace{-10pt}
\end{table}

%% file: table/iso_instant_merge.tex
\begin{table}[t]
  \centering
  \small
  \setlength{\tabcolsep}{6pt}
  \begin{tabular}{l c}
    \toprule
    \textbf{Strategy} & \textbf{Success Rate (\%)} \\
    \midrule
    \textbf{EvoSkill-GUI (Ours)} & \textbf{69.52} \\
    w/o information isolation & 60.95 \\
    w/o instant revision & 62.86 \\
    \bottomrule
  \end{tabular}
  \caption{Ablation of information isolation and instant revision on MobileWorld on Qwen3.6-Plus.}
  \label{tab:isolation_revision_ablation}
  \vspace{-8pt}
\end{table}

%% file: table/meta_info.tex
\begin{table*}[t]
    \centering
    \small
    \setlength{\tabcolsep}{5pt}
    
    \begin{tabularx}{\textwidth}{X c c c c}
        \toprule
        \textbf{Task name} & \textbf{Metadata Score} & \textbf{Full-text Score} & \textbf{Meta@0.6} & \textbf{Full-text@0.6} \\
        \midrule
        Mastodon New Post & 0.8036 & 0.3431 & \cmark & \xmark \\
        Mattermost Create Channel & 1.0000 & 0.4769 & \cmark & \xmark \\
        TaoDian Item Checkout & 1.0000 & 0.6182 & \cmark & \cmark \\
        Mastodon Reply (Moussaka) & 1.0000 & 0.3833 & \cmark & \xmark \\
        Change Wallpaper (Sunflower) & 0.8333 & 0.4184 & \cmark & \xmark \\
        Mattermost Send File & 0.7003 & 0.3414 & \cmark & \xmark \\
        Mattermost Reply Result & 0.7833 & 0.3521 & \cmark & \xmark \\
        Check Cart Price & 0.9500 & 0.4231 & \cmark & \xmark \\
        Mastodon Unfollow & 0.8167 & 0.3876 & \cmark & \xmark \\
        Mastodon Save Photos & 0.8500 & 0.3654 & \cmark & \xmark \\
        Google Maps Alibaba South & 0.9167 & 0.4512 & \cmark & \xmark \\
        Mastodon Conditional Favorite & 0.8500 & 0.2987 & \cmark & \xmark \\
        \midrule
        \textbf{Average / Hit Rate} & \textbf{0.8753} & \textbf{0.4050} & \textbf{12/12} & \textbf{1/12} \\
        \bottomrule
    \end{tabularx}
    \caption{Ablation of metadata for skill retrieval on MobileWorld reuse cases, all cases are completed successfully by EvoSkill-GUI. @0.6 indicate whether the corresponding retrieval score exceeds the threshold $\theta_r=0.6$.}
    \label{tab:metadata_retrieval_ablation}
    
\end{table*}

%% file: table/retrieval.tex
\begin{table}[h]
    \centering
    \small
    \setlength{\tabcolsep}{8pt}
    \begin{tabular}{c c}
        \toprule
        \textbf{Retrieval Threshold $\theta_r$}
        & \textbf{Success Rate (\%)} \\
        \midrule
        0.4 & 55.2 \\
        \textbf{0.6 (Ours)} & \textbf{69.5} \\
        0.8 & 66.7 \\
        \bottomrule
    \end{tabular}
    \caption{Sensitivity of MobileWorld performance to the retrieval threshold $\theta_r$.}
    \label{tab:retrieval_threshold}
    \vspace{-10pt}
\end{table}

%% file: table/reuse.tex
\begin{table}[t]
    \centering
    \small
    \setlength{\tabcolsep}{5pt}
    \begin{tabular}{l c c c}
        \toprule
        \textbf{Skill Source} & \textbf{Tasks} & \textbf{Recovered} & \textbf{Recovery (\%)} \\
        \midrule
        New skills    & 72  & 8/24   & 33.3 \\
        Reused skills & 160 & \textbf{19/65} & \textbf{29.2} \\
        \midrule
        Overall       & 232 & 27/89  & 30.3 \\
        \bottomrule
    \end{tabular}
    \caption{Analysis of skill-reuse on failure-to-success recovery on the AndroidWorld.}
    \label{tab:failure_recovery}
    \vspace{-10pt}
\end{table}

%% file: table/repair_analysis.tex
\begin{table}[t]
\centering
\small
\setlength{\tabcolsep}{4pt}

\begin{tabular}{l l c c}
\toprule
\textbf{Failure Type} & \textbf{Target File} & \textbf{Recovered} & \textbf{Rate} \\
\midrule
Plan-level error & \texttt{plan.md} & 16/23 & 69.6 \\
Grounding error & \texttt{backup.md} & 6/12 & 50.0 \\
Missing contingency & \texttt{recover.md} & 1/4 & 25.0 \\
\bottomrule
\end{tabular}
\caption{Repairability analysis of failed executions by target skill component. }
\label{tab:repairability_analysis}
\vspace{-10pt}
\end{table}

%% file: section/conclusion.tex
\section{Conclusion}
\label{sec:conclusion}

We presented \textbf{EvoSkill-GUI}, a training-free framework that turns deployed GUI skills from static artifacts into living procedural knowledge revisable from execution feedback. By representing each skill as a structured multi-file package and updating it by a reflect-revise-reuse loop with information-isolated critic and tool-restricted edits, EvoSkill-GUI enables a single backbone model to construct, retrieve, execute, and revise skills entirely at inference time. Across MobileWorld, AndroidWorld, and OSWorld, EvoSkill-GUI consistently improves multiple base models without training.

%% file: table/hyperpara.tex
\begin{table}[t]
\centering
\small
\setlength{\tabcolsep}{4pt}
\begin{tabular}{l l l}
\toprule
\textbf{Category} & \textbf{Parameter} & \textbf{Value} \\
\midrule
Execution & Temperature & 0.0 \\
Execution & Max output tokens & 8192 \\
Execution & History screenshots & 3 \\
Execution & Coordinate scale factor & 1000 \\
Execution & Max skill-tool calls/ step & 4 \\
\midrule
Skill generation & Temperature & 0.3 \\
Skill generation & Max output tokens & 4096 \\
\midrule
Critic & Temperature & 0.0 \\
Critic & Max output tokens & 4096 \\
Critic & Max screenshots & 8 \\
Critic & Screenshot max dimension & 1280 \\
\midrule
Retrieval & BM25 $k_1$ & 1.5 \\
Retrieval & BM25 $b$ & 0.75 \\
Retrieval & Keyword threshold & 0.6 \\
Retrieval & Divergence threshold & 0.65 \\
Retrieval & Divergence weight & 0.3 \\
Retrieval & Domain bonus & 0.15 \\
Retrieval & Keyword bonus & $\leq$ 0.2 \\
\midrule
Revision & Revise order & P $\rightarrow$ B $\rightarrow$ C \\
Revision & Failure examples & 3 at most \\
\bottomrule
\end{tabular}
\caption{Main implementation hyperparameters used in EvoSkill-GUI.}
\label{tab:implementation_hparams}
\vspace{-10pt}
\end{table}

%% file: table/token_cost.tex
\begin{table}[h]
    \centering
    \small
    \setlength{\tabcolsep}{3pt}

    \begin{tabular*}{\columnwidth}{
        @{\extracolsep{\fill}} l c c c c @{}
    }
        \toprule
        \textbf{Method}
        & \shortstack{\textbf{Total} \\ \textbf{(M)}}
        & \shortstack{\textbf{Avg./} \\ \textbf{Task (M)}}
        & \shortstack{\textbf{Avg./} \\ \textbf{Round (M)}}
        & \shortstack{\textbf{Acc.} \\ \textbf{(\%)}} \\
        \midrule
        Baseline pass@1
        & 33.20 & 0.284 & 0.284 & 53.3 \\
        \midrule
        Baseline pass@3
        & 103.00 & 0.880 & 0.293 & 62.8 \\
        \midrule
        \shortstack[l]{Single-file skill \\ pass@3}
        & 84.54 & 0.723 & 0.372 & 66.7 \\
        \midrule
        \shortstack[l]{\textbf{EvoSkill-GUI} \\ \textbf{(3 rounds)}}
        & \textbf{94.75}
        & \textbf{0.810}
        & \textbf{0.439}
        & \textbf{69.5} \\
        \bottomrule
    \end{tabular*}

    \caption{Test-time token cost and accuracy on MobileWorld under pass@1 and pass@3 evaluation. M denotes million tokens.}
    \label{tab:token_cost}
    \vspace{-10pt}
\end{table}

%% file: table/a11y_token_time.tex
\begin{table}[h]
\centering
\small
\setlength{\tabcolsep}{4pt}
\begin{tabular}{l c c}
\toprule
\textbf{Task} & \textbf{Token cost(K)} & \textbf{Time(s)} \\
\midrule
AcceptMeetingTask & 67.2 / 71.1 & 437 / 288 \\
MastodonNewPostTask & 79.8 / 79.4 & 476 / 358 \\
SetAlarmTask & 189.1 / 186.0 & 526 / 405 \\
ScheduleLunchViaSmsTask & 800.0 / 820.1 & 1277 / 1125 \\
\bottomrule
\end{tabular}
\caption{Comparison of token usage and execution time with / without accessibility-tree inputs. }
\label{tab:a11y_cost}
\vspace{-8pt}
\end{table}

%% file: main.bbl
\begin{thebibliography}{44}
\providecommand{\natexlab}[1]{#1}

\bibitem[{Allard et~al.(2026)Allard, Teinturier, Xing, and Viaud}]{erl2026}
Marc-Antoine Allard, Arnaud Teinturier, Victor Xing, and Gautier Viaud. 2026.
\newblock \href {https://arxiv.org/abs/2603.24639} {Experiential reflective learning for self-improving llm agents}.
\newblock \emph{Preprint}, arXiv:2603.24639.

\bibitem[{{Anthropic}(2025)}]{anthropic_skill}
{Anthropic}. 2025.
\newblock Agent skills overview.
\newblock \url{https://platform.claude.com/docs/en/agents-and-tools/agent-skills/overview}.
\newblock Accessed: 2026-03-30.

\bibitem[{{Anthropic}(2026)}]{sonnet}
{Anthropic}. 2026.
\newblock Claude sonnet 4.6 system card.
\newblock \url{https://www.anthropic.com/claude-sonnet-4-6-system-card}.

\bibitem[{Bai et~al.(2025)Bai, Cai, Chen, Chen, Chen, Cheng, Deng, Ding, Gao, Ge, Ge, Guo, Huang, Huang, Huang, Hui, Jiang, Li, Li, Li, Li, Lin, Lin, Liu, Liu, Liu, Liu, Liu, Liu, Lu, Luo, Lv, Men, Meng, Ren, Ren, Song, Sun, Tang, Tu, Wan, Wang, Wang, Wang, Wang, Xie, Xu, Xu, Xu, Yang, Yang, Yang, Yang, Yu, Zhang, Zhang, Zhang, Zheng, Zhong, Zhou, Zhou, Zhou, Zhu, and Zhu}]{qwen3vl}
Shuai Bai, Yuxuan Cai, Ruizhe Chen, Keqin Chen, Xionghui Chen, Zesen Cheng, Lianghao Deng, Wei Ding, Chang Gao, Chunjiang Ge, Wenbin Ge, Zhifang Guo, Qidong Huang, Jie Huang, Fei Huang, Binyuan Hui, Shutong Jiang, Zhaohai Li, Mingsheng Li, and 45 others. 2025.
\newblock \href {https://arxiv.org/abs/2511.21631} {Qwen3-vl technical report}.
\newblock \emph{Preprint}, arXiv:2511.21631.

\bibitem[{Chen et~al.(2026)Chen, Li, Solodko, Wang, Jiang, Cui, Hao, Ko, Abdali, Xu, Zheng, Fan, Cameron, Wagle, and Koishida}]{cua_skill}
Tianyi Chen, Yinheng Li, Michael Solodko, Sen Wang, Nan Jiang, Tingyuan Cui, Junheng Hao, Jongwoo Ko, Sara Abdali, Leon Xu, Suzhen Zheng, Hao Fan, Pashmina Cameron, Justin Wagle, and Kazuhito Koishida. 2026.
\newblock \href {https://arxiv.org/abs/2601.21123} {Cua-skill: Develop skills for computer using agent}.
\newblock \emph{Preprint}, arXiv:2601.21123.

\bibitem[{Cheng et~al.(2024)Cheng, Sun, Chu, Xu, Li, Zhang, and Wu}]{screenspot}
Kanzhi Cheng, Qiushi Sun, Yougang Chu, Fangzhi Xu, Yantao Li, Jianbing Zhang, and Zhiyong Wu. 2024.
\newblock \href {https://arxiv.org/abs/2401.10935} {Seeclick: Harnessing gui grounding for advanced visual gui agents}.
\newblock \emph{Preprint}, arXiv:2401.10935.

\bibitem[{Fang et~al.(2026)Fang, Liang, Wang, Wu, Qiao, Xie, Huang, Chen, and Zhang}]{memp2026}
Runnan Fang, Yuan Liang, Xiaobin Wang, Jialong Wu, Shuofei Qiao, Pengjun Xie, Fei Huang, Huajun Chen, and Ningyu Zhang. 2026.
\newblock \href {https://arxiv.org/abs/2508.06433} {Memp: Exploring agent procedural memory}.
\newblock \emph{Preprint}, arXiv:2508.06433.

\bibitem[{Fang et~al.(2025)Fang, Zhang, Zhang, Ma, Yu, Mi, and Yu}]{webevolver}
Tianqing Fang, Hongming Zhang, Zhisong Zhang, Kaixin Ma, Wenhao Yu, Haitao Mi, and Dong Yu. 2025.
\newblock \href {https://arxiv.org/abs/2504.21024} {Webevolver: Enhancing web agent self-improvement with coevolving world model}.
\newblock \emph{Preprint}, arXiv:2504.21024.

\bibitem[{Jiang et~al.(2026)Jiang, Su, Qu, and Fung}]{xskill}
Guanyu Jiang, Zhaochen Su, Xiaoye Qu, and Yi~R. Fung. 2026.
\newblock \href {https://arxiv.org/abs/2603.12056} {Xskill: Continual learning from experience and skills in multimodal agents}.
\newblock \emph{Preprint}, arXiv:2603.12056.

\bibitem[{Kong et~al.(2025)Kong, Zhang, Yang, Gao, Liu, Tong, Cai, Zhou, Zhang, Chen, Liu, Hoi, and Wang}]{mobileworld}
Quyu Kong, Xu~Zhang, Zhenyu Yang, Nolan Gao, Chen Liu, Panrong Tong, Chenglin Cai, Hanzhang Zhou, Jianan Zhang, Liangyu Chen, Zhidan Liu, Steven Hoi, and Yue Wang. 2025.
\newblock \href {https://arxiv.org/abs/2512.19432} {Mobileworld: Benchmarking autonomous mobile agents in agent-user interactive and mcp-augmented environments}.
\newblock \emph{Preprint}, arXiv:2512.19432.

\bibitem[{Li et~al.(2025{\natexlab{a}})Li, Meng, Lin, Luo, Tian, Ma, Huang, and Chua}]{scrrenspotpro}
Kaixin Li, Ziyang Meng, Hongzhan Lin, Ziyang Luo, Yuchen Tian, Jing Ma, Zhiyong Huang, and Tat-Seng Chua. 2025{\natexlab{a}}.
\newblock \href {https://arxiv.org/abs/2504.07981} {Screenspot-pro: Gui grounding for professional high-resolution computer use}.
\newblock \emph{Preprint}, arXiv:2504.07981.

\bibitem[{Li et~al.(2025{\natexlab{b}})Li, Qu, Zhou, Wang, Wen, Du, Lou, Peng, Wang, and Zhang}]{mobileuse}
Ning Li, Xiangmou Qu, Jiamu Zhou, Jun Wang, Muning Wen, Kounianhua Du, Xingyu Lou, Qiuying Peng, Jun Wang, and Weinan Zhang. 2025{\natexlab{b}}.
\newblock \href {https://arxiv.org/abs/2507.16853} {Mobileuse: A gui agent with hierarchical reflection for autonomous mobile operation}.
\newblock \emph{Preprint}, arXiv:2507.16853.

\bibitem[{Lin et~al.(2026)Lin, Liu, Yang, Lyu, Gao, Liu, Lu, Yu, Yang, Li, Ye, and Jiang}]{uivoyager2026}
Zichuan Lin, Feiyu Liu, Yijun Yang, Jiafei Lyu, Yiming Gao, Yicheng Liu, Zhicong Lu, Yangbin Yu, Mingyu Yang, Junyou Li, Deheng Ye, and Jie Jiang. 2026.
\newblock \href {https://arxiv.org/abs/2603.24533} {Ui-voyager: A self-evolving gui agent learning via failed experience}.
\newblock \emph{Preprint}, arXiv:2603.24533.

\bibitem[{Lu et~al.(2025)Lu, Chai, Guo, Yin, Liu, Wang, Xiao, Ren, Xiong, and Li}]{lu2025uir1}
Zhengxi Lu, Yuxiang Chai, Yaxuan Guo, Xi~Yin, Liang Liu, Hao Wang, Han Xiao, Shuai Ren, Guanjing Xiong, and Hongsheng Li. 2025.
\newblock \href {https://arxiv.org/abs/2503.21620} {Ui-r1: Enhancing efficient action prediction of gui agents by reinforcement learning}.
\newblock \emph{Preprint}, arXiv:2503.21620.

\bibitem[{Lu et~al.(2026{\natexlab{a}})Lu, Yao, Han, Wang, Wu, Gu, Cai, Lu, Xiao, Zhuang, and Shen}]{lu2026sdar}
Zhengxi Lu, Zhiyuan Yao, Zhuowen Han, Zi-Han Wang, Jinyang Wu, Qi~Gu, Xunliang Cai, Weiming Lu, Jun Xiao, Yueting Zhuang, and Yongliang Shen. 2026{\natexlab{a}}.
\newblock \href {https://arxiv.org/abs/2605.15155} {Self-distilled agentic reinforcement learning}.
\newblock \emph{Preprint}, arXiv:2605.15155.

\bibitem[{Lu et~al.(2026{\natexlab{b}})Lu, Yao, Wu, Han, Gu, Cai, Lu, Xiao, Zhuang, and Shen}]{skill0}
Zhengxi Lu, Zhiyuan Yao, Jinyang Wu, Chengcheng Han, Qi~Gu, Xunliang Cai, Weiming Lu, Jun Xiao, Yueting Zhuang, and Yongliang Shen. 2026{\natexlab{b}}.
\newblock \href {https://arxiv.org/abs/2604.02268} {Skill0: In-context agentic reinforcement learning for skill internalization}.
\newblock \emph{Preprint}, arXiv:2604.02268.

\bibitem[{Luo et~al.(2026)Luo, Tian, Cao, Luo, Lin, Li, Kong, Yang, and Ma}]{memory_survey2026}
Jinghao Luo, Yuchen Tian, Chuxue Cao, Ziyang Luo, Hongzhan Lin, Kaixin Li, Chuyi Kong, Ruichao Yang, and Jing Ma. 2026.
\newblock \href {https://arxiv.org/abs/2605.06716} {From storage to experience: A survey on the evolution of llm agent memory mechanisms}.
\newblock \emph{Preprint}, arXiv:2605.06716.

\bibitem[{Nguyen et~al.(2025)Nguyen, Chen, Wang, Wu, Park, Hu, Lyu, Wu, Aponte, Xia, Li, Shi, Chen, Lai, Xie, Kim, Zhang, Yu, Tanjim, Ahmed, Mathur, Yoon, Yao, Kveton, Kil, Nguyen, Bui, Zhou, Rossi, and Dernoncourt}]{survey2}
Dang Nguyen, Jian Chen, Yu~Wang, Gang Wu, Namyong Park, Zhengmian Hu, Hanjia Lyu, Junda Wu, Ryan Aponte, Yu~Xia, Xintong Li, Jing Shi, Hongjie Chen, Viet~Dac Lai, Zhouhang Xie, Sungchul Kim, Ruiyi Zhang, Tong Yu, Mehrab Tanjim, and 11 others. 2025.
\newblock \href {https://arxiv.org/abs/2412.13501} {Gui agents: A survey}.
\newblock \emph{Preprint}, arXiv:2412.13501.

\bibitem[{Qin et~al.(2025)Qin, Ye, Fang, Wang, Liang, Tian, Zhang, Li, Li, Huang, Zhong, Li, Yang, Miao, Lin, Liu, Jiang, Ma, Li, Xiao, Cai, Li, Zheng, Jin, Li, Zhou, Wang, Chen, Li, Yang, Liu, Lin, Peng, Liu, and Shi}]{uitars}
Yujia Qin, Yining Ye, Junjie Fang, Haoming Wang, Shihao Liang, Shizuo Tian, Junda Zhang, Jiahao Li, Yunxin Li, Shijue Huang, Wanjun Zhong, Kuanye Li, Jiale Yang, Yu~Miao, Woyu Lin, Longxiang Liu, Xu~Jiang, Qianli Ma, Jingyu Li, and 16 others. 2025.
\newblock \href {https://arxiv.org/abs/2501.12326} {Ui-tars: Pioneering automated gui interaction with native agents}.
\newblock \emph{Preprint}, arXiv:2501.12326.

\bibitem[{{Qwen Team}(2026)}]{qwen3.6-35b-a3b}
{Qwen Team}. 2026.
\newblock \href {https://qwen.ai/blog?id=qwen3.6-35b-a3b} {{Qwen3.6-35B-A3B}: Agentic coding power, now open to all}.

\bibitem[{Rawles et~al.(2025)Rawles, Clinckemaillie, Chang, Waltz, Lau, Fair, Li, Bishop, Li, Campbell-Ajala, Toyama, Berry, Tyamagundlu, Lillicrap, and Riva}]{androidworld}
Christopher Rawles, Sarah Clinckemaillie, Yifan Chang, Jonathan Waltz, Gabrielle Lau, Marybeth Fair, Alice Li, William Bishop, Wei Li, Folawiyo Campbell-Ajala, Daniel Toyama, Robert Berry, Divya Tyamagundlu, Timothy Lillicrap, and Oriana Riva. 2025.
\newblock \href {https://arxiv.org/abs/2405.14573} {Androidworld: A dynamic benchmarking environment for autonomous agents}.
\newblock \emph{Preprint}, arXiv:2405.14573.

\bibitem[{Sun et~al.(2025)Sun, Liu, Zang, Cao, Dong, Wu, Lin, and Wang}]{seagent}
Zeyi Sun, Ziyu Liu, Yuhang Zang, Yuhang Cao, Xiaoyi Dong, Tong Wu, Dahua Lin, and Jiaqi Wang. 2025.
\newblock \href {https://arxiv.org/abs/2508.04700} {Seagent: Self-evolving computer use agent with autonomous learning from experience}.
\newblock \emph{Preprint}, arXiv:2508.04700.

\bibitem[{Tang et~al.(2025{\natexlab{a}})Tang, Gu, Lu, Liu, Shen, Meng, Wang, Zhang, Shen, Lu, Xiao, and Zhuang}]{tang2025guig2}
Fei Tang, Zhangxuan Gu, Zhengxi Lu, Xuyang Liu, Shuheng Shen, Changhua Meng, Wen Wang, Wenqi Zhang, Yongliang Shen, Weiming Lu, Jun Xiao, and Yueting Zhuang. 2025{\natexlab{a}}.
\newblock \href {https://arxiv.org/abs/2507.15846} {Gui-g$^2$: Gaussian reward modeling for gui grounding}.
\newblock \emph{Preprint}, arXiv:2507.15846.

\bibitem[{Tang et~al.(2025{\natexlab{b}})Tang, Xu, Zhang, Chen, Wu, Shen, Zhang, Hou, Tan, Yan, Song, Shao, Lu, Xiao, and Zhuang}]{survey1}
Fei Tang, Haolei Xu, Hang Zhang, Siqi Chen, Xingyu Wu, Yongliang Shen, Wenqi Zhang, Guiyang Hou, Zeqi Tan, Yuchen Yan, Kaitao Song, Jian Shao, Weiming Lu, Jun Xiao, and Yueting Zhuang. 2025{\natexlab{b}}.
\newblock \href {https://arxiv.org/abs/2504.13865} {A survey on (m)llm-based gui agents}.
\newblock \emph{Preprint}, arXiv:2504.13865.

\bibitem[{Tu et~al.(2026)Tu, Xu, Zhang, Zhang, Lan, Li, and Zhao}]{d2skill}
Songjun Tu, Chengdong Xu, Qichao Zhang, Yaocheng Zhang, Xiangyuan Lan, Linjing Li, and Dongbin Zhao. 2026.
\newblock \href {https://arxiv.org/abs/2603.28716} {Dynamic dual-granularity skill bank for agentic rl}.
\newblock \emph{Preprint}, arXiv:2603.28716.

\bibitem[{Wang et~al.(2025{\natexlab{a}})Wang, Wang, Lu, Yang, Xie, Wang, Deng, Guo, Xu, Wu, Shen, Li, Li, Li, Chen, Zheng, Li, Lei, Cao, Fu, Shin, Shin, Hu, Wang, Chen, Ye, Zhang, Du, Hu, Chen, Zhou, Yao, Chen, Gu, Wang, Wang, Yang, Zhong, Sung, Charles, Yang, and Yu}]{opencua}
Xinyuan Wang, Bowen Wang, Dunjie Lu, Junlin Yang, Tianbao Xie, Junli Wang, Jiaqi Deng, Xiaole Guo, Yiheng Xu, Chen~Henry Wu, Zhennan Shen, Zhuokai Li, Ryan Li, Xiaochuan Li, Junda Chen, Boyuan Zheng, Peihang Li, Fangyu Lei, Ruisheng Cao, and 23 others. 2025{\natexlab{a}}.
\newblock \href {https://arxiv.org/abs/2508.09123} {Opencua: Open foundations for computer-use agents}.
\newblock \emph{Preprint}, arXiv:2508.09123.

\bibitem[{Wang et~al.(2026)Wang, Wu, Zhang, Zhang, Yao, Faisal, Peng, Qin, Nath, Lin, Bansal, Zhang, Rajmohan, Gao, and Yao}]{webxskill}
Zhaoyang Wang, Qianhui Wu, Xuchao Zhang, Chaoyun Zhang, Wenlin Yao, Fazle~Elahi Faisal, Baolin Peng, Si~Qin, Suman Nath, Qingwei Lin, Chetan Bansal, Dongmei Zhang, Saravan Rajmohan, Jianfeng Gao, and Huaxiu Yao. 2026.
\newblock \href {https://arxiv.org/abs/2604.13318} {Webxskill: Skill learning for autonomous web agents}.
\newblock \emph{Preprint}, arXiv:2604.13318.

\bibitem[{Wang et~al.(2025{\natexlab{b}})Wang, Xu, Wang, Zhang, Yan, Zhang, Huang, and Ji}]{mobile_agent_e}
Zhenhailong Wang, Haiyang Xu, Junyang Wang, Xi~Zhang, Ming Yan, Ji~Zhang, Fei Huang, and Heng Ji. 2025{\natexlab{b}}.
\newblock \href {https://arxiv.org/abs/2501.11733} {Mobile-agent-e: Self-evolving mobile assistant for complex tasks}.
\newblock \emph{Preprint}, arXiv:2501.11733.

\bibitem[{Wu et~al.(2025)Wu, Ma, Wang, Yu, Lu, and Liu}]{gui_reflection}
Penghao Wu, Shengnan Ma, Bo~Wang, Jiaheng Yu, Lewei Lu, and Ziwei Liu. 2025.
\newblock \href {https://arxiv.org/abs/2506.08012} {Gui-reflection: Empowering multimodal gui models with self-reflection behavior}.
\newblock \emph{Preprint}, arXiv:2506.08012.

\bibitem[{Xia et~al.(2026)Xia, Chen, Wang, Liu, Zeng, Wang, Han, Zhou, Zhao, Chen, Zheng, Xie, and Yao}]{skillrl}
Peng Xia, Jianwen Chen, Hanyang Wang, Jiaqi Liu, Kaide Zeng, Yu~Wang, Siwei Han, Yiyang Zhou, Xujiang Zhao, Haifeng Chen, Zeyu Zheng, Cihang Xie, and Huaxiu Yao. 2026.
\newblock \href {https://arxiv.org/abs/2602.08234} {Skillrl: Evolving agents via recursive skill-augmented reinforcement learning}.
\newblock \emph{Preprint}, arXiv:2602.08234.

\bibitem[{Xiao et~al.(2026)Xiao, Wang, Wang, Liu, Chai, Pan, Zhou, Chen, Wen, and Li}]{ui_mem}
Han Xiao, Guozhi Wang, Hao Wang, Shilong Liu, Yuxiang Chai, Yue Pan, Yufeng Zhou, Xiaoxin Chen, Yafei Wen, and Hongsheng Li. 2026.
\newblock \href {https://arxiv.org/abs/2602.05832} {Ui-mem: Self-evolving experience memory for online reinforcement learning in mobile gui agents}.
\newblock \emph{Preprint}, arXiv:2602.05832.

\bibitem[{Xie et~al.(2024)Xie, Zhang, Chen, Li, Zhao, Cao, Hua, Cheng, Shin, Lei, Liu, Xu, Zhou, Savarese, Xiong, Zhong, and Yu}]{osworld}
Tianbao Xie, Danyang Zhang, Jixuan Chen, Xiaochuan Li, Siheng Zhao, Ruisheng Cao, Toh~Jing Hua, Zhoujun Cheng, Dongchan Shin, Fangyu Lei, Yitao Liu, Yiheng Xu, Shuyan Zhou, Silvio Savarese, Caiming Xiong, Victor Zhong, and Tao Yu. 2024.
\newblock \href {https://arxiv.org/abs/2404.07972} {Osworld: Benchmarking multimodal agents for open-ended tasks in real computer environments}.
\newblock \emph{Preprint}, arXiv:2404.07972.

\bibitem[{Xu and Yan(2026)}]{survey_skills2026}
Renjun Xu and Yang Yan. 2026.
\newblock \href {https://arxiv.org/abs/2602.12430} {Agent skills for large language models: Architecture, acquisition, security, and the path forward}.
\newblock \emph{Preprint}, arXiv:2602.12430.

\bibitem[{Xue et~al.(2026)Xue, Peng, Huang, Guo, Han, Wang, Wang, Zhang, Yang, Zhao, Ding, Ma, Xie, Pei, Cai, and Qiu}]{evocua}
Taofeng Xue, Chong Peng, Mianqiu Huang, Linsen Guo, Tiancheng Han, Haozhe Wang, Jianing Wang, Xiaocheng Zhang, Xin Yang, Dengchang Zhao, Jinrui Ding, Xiandi Ma, Yuchen Xie, Peng Pei, Xunliang Cai, and Xipeng Qiu. 2026.
\newblock \href {https://arxiv.org/abs/2601.15876} {Evocua: Evolving computer use agents via learning from scalable synthetic experience}.
\newblock \emph{Preprint}, arXiv:2601.15876.

\bibitem[{Ye et~al.(2025{\natexlab{a}})Ye, Zhang, Xu, Liu, Wang, Zhu, Zheng, Gao, Cao, Lu, Liao, Zheng, Huang, Zhou, and Yan}]{ye2025mobileagentv3}
Jiabo Ye, Xi~Zhang, Haiyang Xu, Haowei Liu, Junyang Wang, Zhaoqing Zhu, Ziwei Zheng, Feiyu Gao, Junjie Cao, Zhengxi Lu, Jitong Liao, Qi~Zheng, Fei Huang, Jingren Zhou, and Ming Yan. 2025{\natexlab{a}}.
\newblock \href {https://arxiv.org/abs/2508.15144} {Mobile-agent-v3: Fundamental agents for gui automation}.
\newblock \emph{Preprint}, arXiv:2508.15144.

\bibitem[{Ye et~al.(2025{\natexlab{b}})Ye, Zhang, Xu, Liu, Wang, Zhu, Zheng, Gao, Cao, Lu, Liao, Zheng, Huang, Zhou, and Yan}]{gui_owl}
Jiabo Ye, Xi~Zhang, Haiyang Xu, Haowei Liu, Junyang Wang, Zhaoqing Zhu, Ziwei Zheng, Feiyu Gao, Junjie Cao, Zhengxi Lu, Jitong Liao, Qi~Zheng, Fei Huang, Jingren Zhou, and Ming Yan. 2025{\natexlab{b}}.
\newblock \href {https://arxiv.org/abs/2508.15144} {Mobile-agent-v3: Fundamental agents for gui automation}.
\newblock \emph{Preprint}, arXiv:2508.15144.

\bibitem[{Zhang et~al.(2026{\natexlab{a}})Zhang, Fan, Zou, Chen, Wang, Zhou, Li, Huang, Yao, Zheng, Liu, Li, and Yu}]{coevoskills}
Hanrong Zhang, Shicheng Fan, Henry~Peng Zou, Yankai Chen, Zhenting Wang, Jiayu Zhou, Chengze Li, Wei-Chieh Huang, Yifei Yao, Kening Zheng, Xue Liu, Xiaoxiao Li, and Philip~S. Yu. 2026{\natexlab{a}}.
\newblock \href {https://arxiv.org/abs/2604.01687} {Coevoskills: Self-evolving agent skills via co-evolutionary verification}.
\newblock \emph{Preprint}, arXiv:2604.01687.

\bibitem[{Zhang et~al.(2026{\natexlab{b}})Zhang, Shao, Li, Lin, Fu, Wang, Jiao, Lu, Liu, Zhang, and Yu}]{mmskills}
Kangning Zhang, Shuai Shao, Qingyao Li, Jianghao Lin, Lingyue Fu, Shijian Wang, Wenxiang Jiao, Yuan Lu, Weiwen Liu, Weinan Zhang, and Yong Yu. 2026{\natexlab{b}}.
\newblock \href {https://arxiv.org/abs/2605.13527} {Mmskills: Towards multimodal skills for general visual agents}.
\newblock \emph{Preprint}, arXiv:2605.13527.

\bibitem[{Zhang et~al.(2026{\natexlab{c}})Zhang, Xue, Wu, Chen, Liu, He, Shao, Liu, Xu, Pan, and Wang}]{verigui2026}
Yuzhe Zhang, Xianwei Xue, Xingyong Wu, Mengke Chen, Chen Liu, Xinran He, Run Shao, Feiran Liu, Huanmin Xu, Qiutong Pan, and Haiwei Wang. 2026{\natexlab{c}}.
\newblock \href {https://arxiv.org/abs/2604.05477} {Don't act blindly: Robust gui automation via action-effect verification and self-correction}.
\newblock \emph{Preprint}, arXiv:2604.05477.

\bibitem[{Zhang et~al.(2025)Zhang, Liu, Zhang, Wang, Chen, and Lu}]{ui_evol}
Ziyun Zhang, Xinyi Liu, Xiaoyi Zhang, Jun Wang, Gang Chen, and Yan Lu. 2025.
\newblock \href {https://arxiv.org/abs/2505.21964} {Ui-evol: Automatic knowledge evolving for computer use agents}.
\newblock \emph{Preprint}, arXiv:2505.21964.

\bibitem[{Zhao et~al.(2026)Zhao, Wang, Zhang, Yao, and Wang}]{hy_mem}
Xiaochen Zhao, Kaikai Wang, Xiaowen Zhang, Chen Yao, and Aili Wang. 2026.
\newblock \href {https://arxiv.org/abs/2602.13933} {Hymem: Hybrid memory architecture with dynamic retrieval scheduling}.
\newblock \emph{Preprint}, arXiv:2602.13933.

\bibitem[{Zhou et~al.(2025)Zhou, Zhang, Tong, Zhang, Chen, Kong, Cai, Liu, Wang, Zhou, and Hoi}]{mai_ui}
Hanzhang Zhou, Xu~Zhang, Panrong Tong, Jianan Zhang, Liangyu Chen, Quyu Kong, Chenglin Cai, Chen Liu, Yue Wang, Jingren Zhou, and Steven Hoi. 2025.
\newblock \href {https://arxiv.org/abs/2512.22047} {Mai-ui technical report: Real-world centric foundation gui agents}.
\newblock \emph{Preprint}, arXiv:2512.22047.

\bibitem[{Zhou et~al.(2026)Zhou, Shu, Su, Du, Fang, and Lin}]{surveyagentskills}
Yingli Zhou, Wang Shu, Yaodong Su, Wenchuan Du, Yixiang Fang, and Xuemin Lin. 2026.
\newblock \href {https://arxiv.org/abs/2605.07358} {A comprehensive survey on agent skills: Taxonomy, techniques, and applications}.
\newblock \emph{Preprint}, arXiv:2605.07358.

\bibitem[{Zhu et~al.(2026)Zhu, Wu, Zhou, Wang, and Huang}]{hymem2026}
Sibo Zhu, Wenyi Wu, Kun Zhou, Stephen Wang, and Biwei Huang. 2026.
\newblock \href {https://arxiv.org/abs/2603.10291} {Hybrid self-evolving structured memory for gui agents}.
\newblock \emph{Preprint}, arXiv:2603.10291.

\end{thebibliography}
